\documentclass[10pt,twocolumn,letterpaper]{article}
\usepackage{cvpr}              

\usepackage[dvipsnames]{xcolor}
\definecolor{cvprblue}{rgb}{0.21,0.49,0.74}
\usepackage[pagebackref,breaklinks,colorlinks,citecolor=cvprblue]{hyperref}
\usepackage{url}
\usepackage{graphicx}
\usepackage{caption}
\usepackage{subcaption}
\usepackage{amsmath,amsthm,amssymb,amsfonts,bm}
\usepackage[ruled,noline]{algorithm2e}
\usepackage{wrapfig}
\usepackage{makecell}
\usepackage[accsupp]{axessibility} 

\graphicspath{{./media/}}

\def\eg{\emph{e.g., }}
\def\ie{\emph{i.e., }}

\def\ee{\mathbb{E}}      

\def\vv#1{\bm{#1}}  
\def\mm#1{\bm{#1}}  
\def\rn#1{\textnormal{#1}}  
\def\rv#1{\mathbf{#1}}     

\def\pp#1{\left( #1 \right) }                 
\def\ppi#1{( #1 ) }                           
\def\pb#1{\left[ #1 \right] }                 
\def\pbi#1{[ #1 ] }                           
\def\br#1{\left\lbrace #1 \right\rbrace }     
\def\bri#1{\lbrace #1 \rbrace }               

\def\softmax{\mathrm{Softmax} }

\def\pr#1{p_{#1}}  

\def\paperID{12640}
\def\confName{CVPR}
\def\confYear{2024}

\title{Uncertainty Visualization via Low-Dimensional Posterior Projections}

\author{
Omer Yair,
Elias Nehme \&
Tomer Michaeli \\
Technion - Israel Institute of Technology\\
Haifa, Israel\\
{\tt\small omeryair@gmail.com},
{\tt\small seliasne@gmail.com},
{\tt\small tomer.m@ee.technion.ac.il} \\
}

\begin{document}

\maketitle


\begin{abstract}

In ill-posed inverse problems, it is commonly desirable to obtain insight into the full spectrum of plausible solutions, rather than extracting only a single reconstruction. Information about the plausible solutions and their likelihoods is encoded in the posterior distribution. However, for high-dimensional data, this distribution is challenging to visualize. 
In this work, we introduce a new approach for estimating and visualizing posteriors by employing energy-based models (EBMs) over low-dimensional subspaces. Specifically, we train a conditional EBM that receives an input measurement and a set of directions that span some low-dimensional subspace of solutions, and outputs the probability density function of the posterior within that space. We demonstrate the effectiveness of our method across a diverse range of datasets and image restoration problems, showcasing its strength in uncertainty quantification and visualization. As we show, our method outperforms a baseline that projects samples from a diffusion-based posterior sampler, while being orders of magnitude faster. Furthermore, it is more accurate than a baseline that assumes a Gaussian posterior. Code is available at \url{https://github.com/yairomer/PPDE}

\end{abstract}

\section{Introduction}

\begin{figure}[ht!]
    \centering
    \includegraphics[width=0.9\linewidth]{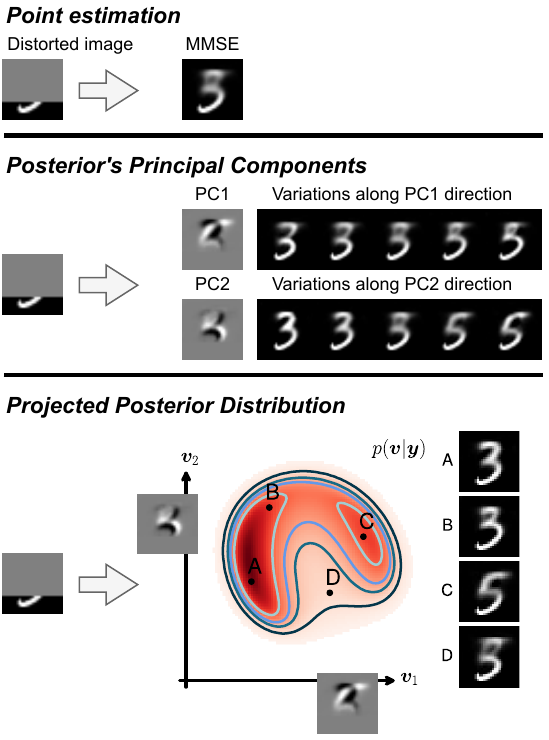}
    \caption{Informed uncertainty visualization. Point estimation methods receive a distorted image and output only a single solution, \eg the MMSE estimator (top). NPPC \citep{nehme2023uncertainty} complements MMSE estimators with input-adaptive uncertainty directions (principal components of the posterior) without modeling output likelihood (middle). Our method (bottom) learns the input-adaptive projected posterior distribution, facilitating a likelihood-informed uncertainty visualization.}
    \label{fig:concept}
\end{figure}

Interpreting and communicating prediction uncertainty plays a key role in advancing trustworthy models that could assist decision-makers. In the context of imaging, many practical problems are ill-posed, so that a range of plausible explanations exist for any given input. In such cases, it is beneficial to provide the user with tools to efficiently explore and visualize the set of all admissible solutions. This is especially crucial in safety-critical domains such as scientific and medical image analysis \citep{ounkomol2018label, christiansen2018silico, rivenson2019virtual, falk2019u}, where mistaken predictions could influence human life. 

The information about the plausible solutions and their likelihoods is fully encoded in the posterior distribution. However, high-dimensional posteriors are hard to estimate and practically impossible to visualize. One way to visualize uncertainty is to settle with generating samples from the posterior \citep{song2021solving,ohayon2021high,kawar2022denoising,wang2023ddnm,chung2022diffusion,li2022mat,lugmayr2022repaint, saharia2022image, saharia2022palette, bendel2022regularized}. However, navigating many samples per input is highly inefficient and often an impractical way of communicating uncertainty \cite{cohen2023posterior}. Indeed, in complex domains and high levels of uncertainty, users may need to examine hundreds of posterior samples to confidently confirm or refute their suspicion about the unobserved ground-truth image. Several works proposed to manipulate the sampling process to produce a small set of meaningful samples highlighting posterior diversity \citep{mao2019mode, yu2020inclusive, sehwag2022generating, cohen2023posterior}. Nonetheless, methods that are based on state-of-the-art (diffusion based) samplers are unacceptably slow due to their iterative sampling process. 

Another way to summarize uncertainty in image restoration problems is via per-pixel heatmaps \citep{kendall2017uncertainties,amini2020deep,wang2019aleatoric,angelopoulos2022image,horwitz2022conffusion}. However, such maps ignore the correlations between pixels in the recovered image and thus typically result in non-semantic and unnecessarily inflated uncertainty estimates that are of limited practical use.

Recently, several works proposed to visualize posterior uncertainty by traversing along the principal components (PCs) of the posterior \citep{nehme2023uncertainty,belhasin2023principal}. These visualizations shed light on the main modes of variation along which the solution could vary given the input, and have thus been demonstrated to be natural for signals with strong correlations, like images. These methods are also mathematically appealing since one-dimensional statistics of the projected posterior along the PCs (like variances or quantiles) provide a tighter and more accurate description of the underlying uncertainty~\citep{belhasin2023principal}. Nonetheless, methods in this category do not provide estimates for the likelihoods of the projected solutions within the low-dimensional subspace. This hinders the user's ability to interpret and visualize the effective support of the projected posterior. 

In this work, we propose to model the projected posterior using a conditional energy-based model (EBM) that receives a degraded measurement and an affine subspace (parameterized by an origin point and a set of directions) and can output the projection of the posterior probability for any queried point within that subspace. This allows us to visualize the projected posterior within any low-dimensional space (\eg the one spanned by the dominant posterior PCs), facilitating informed navigation of posterior uncertainty in both one and two dimensions (Fig.~\ref{fig:concept}). Our method, which we coin \emph{projected posterior distribution estimation} (PPDE), is a general technique for uncertainty visualization that is seamlessly transferable across tasks and datasets. We demonstrate it with the affine subspace that passes through the point prediction provided by an MSE-optimized model and spanned by the posterior PCs provided by the NPPC method \citep{nehme2023uncertainty} and . 
However, our method is not constrained to this specific choice and can wrap around any point estimate and an accompanying set of directions.

As we illustrate, our proposed visualization approach cannot be achieved with methods that explicitly learn the high-dimensional posterior distribution, like conditional invertible models (\eg SRFlow~\citep{lugmayr2020srflow}) or score-based models that enable the calculation of the exact likelihood using the probability flow ODE \citep{song2020score,song2021maximum}. In particular, the naive approach of evaluating and plotting the posterior along a 2D or 1D slice through the high-dimensional distribution gives meaningless results, as it will almost surely miss the high-probability regions in space (see Fig.~\ref{fig:proj_vs_slice} and the explanation in Sec.~\ref{sec:proj}). By contrast, our approach corresponds to \emph{projecting} the distribution onto the 2D or 1D slice (\ie integrating over all other dimensions), an effect that is practically impossible to achieve with models that output the density in the high-dimensional space. 

We demonstrate the practical benefit of our method on multiple inverse problems in imaging. Our learned posteriors are quantitatively compared to a Gaussian approximation obtained from \cite{nehme2023uncertainty}, as well as to a baseline that projects samples generated by a diffusion-based posterior sampler and applies kernel-density estimation (KDE) in the low-dimensional space. In both cases, we show a significant improvement in sample log-likelihood across tasks and datasets. This illustrates the benefit offered by our method compared to works that approximate the posterior with a Gaussian \citep{dorta2018structured,monteiro2020stochastic,meng2021estimating,nussbaum2022structuring} or attempt to rely on posterior samplers \cite{belhasin2023principal}, setting forth a new approach to proper uncertainty visualization.


\section{Related Work}

In the deep learning literature (\eg \citep{gal2016dropout,kendall2017uncertainties}), predictive uncertainty is often decomposed into two main sources: (i)~epistemic/model uncertainty \citep{neal2012bayesian, salimans2015markov, blundell2015weight, lakshminarayanan2017simple, louizos2017multiplicative, ritter2018scalable, pearce2018high, malinin2018predictive, izmailov2020subspace}, which stems from imperfect knowledge of model parameters and can be reduced by acquiring additional training data, and (ii) aleatoric/data uncertainty \citep{kendall2017uncertainties,wang2019aleatoric,ohayon2021high,angelopoulos2022image,kawar2022denoising,wang2023ddnm}, which is inherent to the task (\eg due to the degradation and measurement noise), and cannot be reduced even with infinite data. We focus on the latter.

Until recently, the majority of methods for uncertainty quantification in imaging problems focused on per-pixel estimates \eg in the form of variance heatmaps \citep{kendall2017uncertainties} or confidence intervals \citep{angelopoulos2022image}. However, these techniques ignore inter-pixel correlations, which are very important in visual data. Distribution-free methods such as risk-controlling prediction sets (RCPS) \citep{bates2021distribution} treat all pixels jointly by aggregating and controlling their risk (\eg in image segmentation). 
However, RCPS only provides upper and lower bounds on the solution set without the ability to navigate within-set possibilities, and it also requires an extra data split which may not be readily available in data-scarce settings. A follow-up work \citep{sankaranarayanansemantic}, employed a similar idea in the latent space of StyleGAN, demonstrating some informative visualizations of uncertainty. However, this work is limited in its ability to treat real images and it requires previously identified disentangled latent directions.

Another alternative for exploring data uncertainty is posterior sampling using conditional generative models such as \citep{lugmayr2020srflow, ohayon2021high, song2021solving, kawar2021snips, kawar2022denoising, chung2022diffusion, li2022mat, lugmayr2022repaint, saharia2022palette, saharia2022image, bendel2022regularized, wang2023ddnm, chung2023fast}, with score-based/diffusion models \citep{hyvarinen2005estimation, sohl2015deep, song2019generative, ho2020denoising} pulling ahead in the last two years. Posterior sampling can in principle offer a large solution set for any given input, which can, in turn, be summarized to useful uncertainty estimates \eg using PCA \citep{belhasin2023principal}. However, this strategy is extremely slow with unbearable run times for modern state-of-the-art models, despite promising recent efforts to speed them up \citep{salimans2022progressive, luhman2021knowledge, meng2022on, song2023consistency}. 

Some works proposed to approximate posteriors with more simple distributions that account for pixel correlations (\eg correlated Gaussians \citep{dorta2018structured, monteiro2020stochastic, meng2021estimating, nussbaum2022structuring}), which are fast to infer. Most recently, two methods were proposed to output the top PCs of the posterior distribution directly without imposing a distributional assumption \citep{nehme2023uncertainty,manor2023posterior}. Specifically, NPPC \citep{nehme2023uncertainty} was shown to provide useful uncertainty estimates while being extremely fast. Here we use it as a building block to demonstrate our method.

A visualization of projected posterior distributions, as we propose here, can theoretically be achieved by generating many posterior samples with some stochastic inverse-problem solver (\eg \citep{wang2023ddnm}), projecting them onto the desired subspace, and estimating the distribution of the low-dimensional projections using \eg kernel density estimation (KDE). However, this approach is impractical for real-world applications as the computational burden of generating a sufficient number of samples is very high with state-of-the-art posterior samplers. Furthermore, as we show, our method outperforms this approach in terms of average negative log-likelihood (NLL) even when using a large sample.

Another way to obtain an approximation of the projected posterior was recently proposed in the concurrent work of \citep{manor2023posterior}. Specifically, this work presents a training-free method for estimating the one-dimensional projected posterior distribution along any desired direction in the task of Gaussian denoising. Our approach offers two key advantages over this work: (i) It is applicable to arbitrary inverse problems and not only to Gaussian denoising; (ii) It is applicable to arbitrary subspace dimensions (\eg 2D) and not constrained to only 1D  projections.


\section{Method}

\begin{figure}
    \centering
    \includegraphics[width=\linewidth]{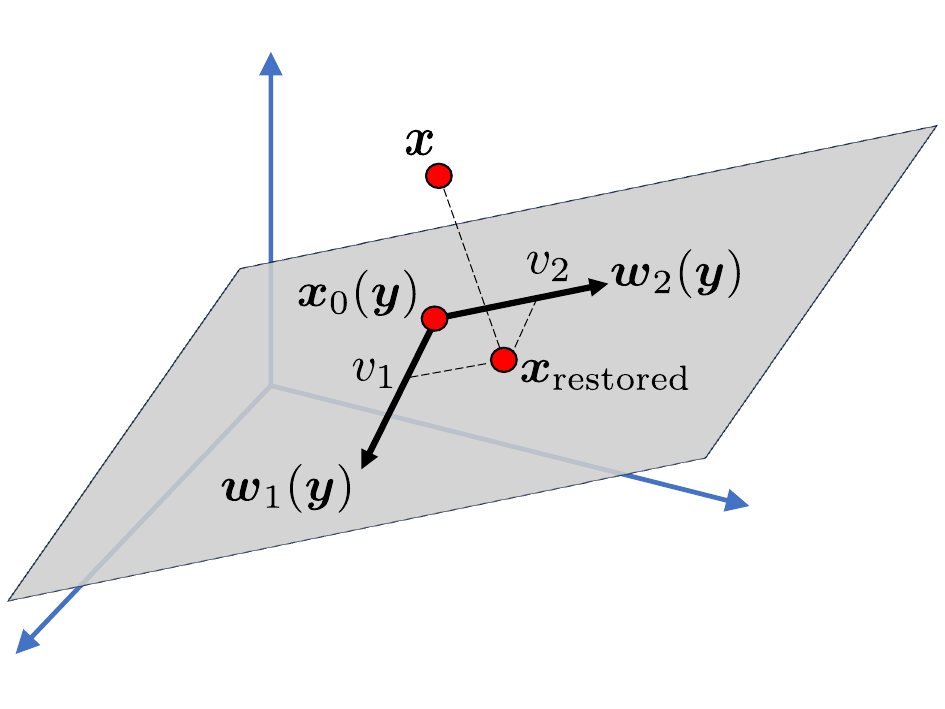}
    \caption{Notations illustration. An image $\bm{x}$ is projected onto a measurement-adaptive affine subspace $\mathcal{A}(\vv{y})$, and represented by its projection coefficients $\vv{v}=[v_1,v_2]^{\top}$. The restored image within the subspace is denoted by $\vv{x}_{\text{restored}}$.}\label{fig:subspaceA}
\end{figure}

\subsection{Projecting the posterior distribution}\label{sec:proj}

Our goal is to visualize the uncertainty when predicting a signal $\vv{x}$ based on measurements $\vv{y}$. In the context of imaging, $\vv{y}$ often represents a degraded version of $\vv{x}$ (\eg noisy, blurry). We assume that $\vv{x}$ and $\vv{y}$ are realizations of random vectors $\rv{x}$ and $\rv{y}$, respectively. The desired uncertainty we aim to explore is therefore encapsulated in the posterior distribution $\pr{\rv{x}|\rv{y}}(\vv{x}|\vv{y})$. As mentioned earlier, this distribution is typically defined over a high-dimensional space, and therefore visualizing it requires projection onto a lower-dimensional manifold (\eg 1D or 2D). 

It is important to note that the manifold that best suits uncertainty visualization may generally depend on the observation $\vv{y}$.  Here we choose the manifold to be an input-dependent affine subspace of the form 
\begin{equation}
\left\{\vv{x}:\vv{x}=\vv{x}_0(\vv{y})+\mm{W}(\vv{y})\vv{v}, \;\vv{v}\in\mathbb{R}^K\right\},
\end{equation}
where $\vv{x}_0(\vv{y})$ is an origin point and $\mm{W}(\vv{y})$ is a matrix with $K$ orthonormal columns, 
\begin{equation}
\mm{W}(\vv{y})=\begin{bmatrix}| & | &  & | \\ \vv{w}_1(\vv{y}) & \vv{w}_2(\vv{y}) & \dots & \vv{w}_K(\vv{y}) \\ | & | &  & | \end{bmatrix}.
\end{equation}
For conciceness, we denote the pair $\{\vv{x}_0(\vv{y}),\mm{W}(\vv{y})\}$ by
\begin{equation}
\mathcal{A}(\vv{y})=\br{\vv{x}_0(\vv{y}),W(\vv{y})}
\end{equation}
and with slight abuse of terminology, refer to $\mathcal{A}(\vv{y})$ as our affine subspace (or just subspace). Our method can be used along with any method of producing a subspace $\mathcal{A}(\vv{y})$ for a given input $\vv{y}$. Here we choose $\vv{x}_0\ppi{\vv{y}}$ to be a minimum MSE (MMSE) predictor, and use the NPPC method \cite{nehme2023uncertainty} for producing $\mm{W}\ppi{\vv{y}}$ (see Sec.~\ref{subsec:selecting_subspace}).

For a given input $\vv{y}$, our goal is to visualize the posterior distribution of the coordinates of $\rv{x}$ projected onto the subspace $\mathcal{A}(\vv{y})$, which we denote by
\begin{equation}
\rv{v} = \mm{W}\ppi{\vv{y}}^\top\ppi{\rv{x}-\vv{x}_0\ppi{\vv{y}}}.
\end{equation}
Namely, we are interested in the \emph{projected posterior distribution} (PPD) $\pr{\rv{v}|\rv{y}}(\vv{v}|\vv{y})$.  When exploring this distribution, a user can visually inspect any particular $\vv{v}$ by projecting it back into pixel space as
\begin{equation}
\vv{x}_{\text{restored}} = \vv{x}_0(\vv{y})+\mm{W}(\vv{y})\vv{v}.
\end{equation}
Figure~\ref{fig:subspaceA} summarizes our notations.

\begin{figure}
    \centering
    \begin{subfigure}[b]{\linewidth}\centering
        \includegraphics[width=0.7\linewidth]{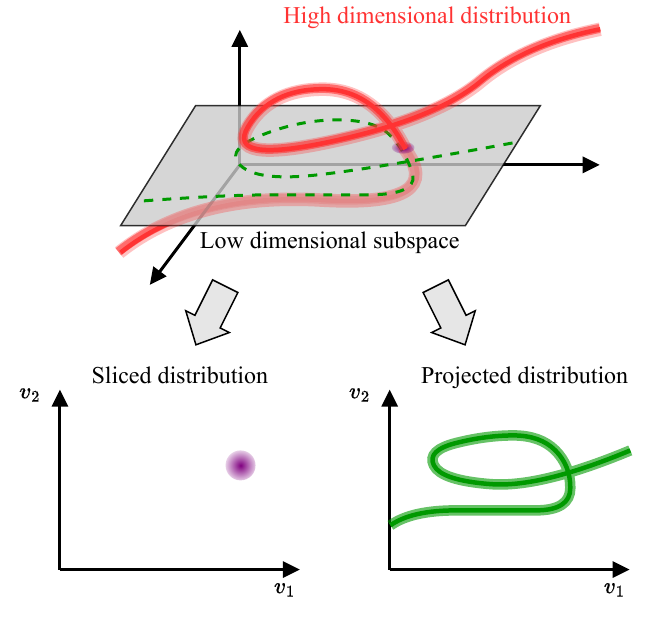}
        \caption{Schematic illustration.}
    \end{subfigure}
    
    \begin{subfigure}[b]{\linewidth}\centering
        \includegraphics[width=0.95\linewidth]{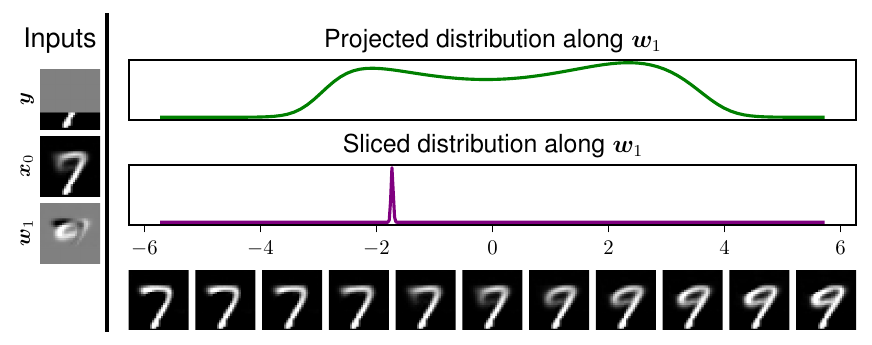}
        \caption{Digit inpainting.}
    \end{subfigure}
    \caption{Posterior slicing vs. projection. (a) Schematic illustrating the difference between the slice and the projection of a high-dimensional manifold onto a low-dimensional subspace. (b) Sliced and projected posterior comparison for a 1D affine subspace defined by the posterior mean and the first PC in the task of digit inpainting. The sliced posterior has been computed using an EBM.}\label{fig:proj_vs_slice}
\end{figure}

It is instructive to note the difference between the PPD and a sliced posterior distribution (\ie the high-dimensional posterior evaluated on $\mathcal{A}(\vv{y})$). Any $\vv{x}$ can be expressed in terms of its projection onto $\mathcal{A}(\vv{y})$ and its projection onto the orthogonal complement of $\mathcal{A}(\vv{y})$ as  $\vv{x}=\vv{x}_0(\vv{y})+\mm{W}(\vv{y})\vv{v}+\mm{W}^\perp(\vv{y})\vv{u}$. Here, $\mm{W}^\perp(\vv{y})$ is a matrix with orthonormal columns that span $\text{Range}^\perp\{\mm{W}(\vv{y})\}$, and $\vv{u}$ corresponds to the coordinates within that subspace. Now, the sliced posterior distribution corresponds to evaluating the posterior at $\vv{u}=0$ in this decomposition. Namely, it is given by\footnote{The function $f^{\text{sliced}}$ is not a density function as it is not normalized.}
\begin{align}
    f^{\text{sliced}}(\vv{v}|\vv{y})=\pr{\rv{x}|\rv{y}}\left(\vv{x}_0(\vv{y})+\mm{W}(\vv{y})\vv{v}\middle|\vv{y}\right).
\end{align}
In contrast, the PPD at a point $\vv{v}$ corresponds to the integral of the posterior over all points whose projection onto $\mathcal{A}(\vv{y})$ has coordinates $\vv{v}$. In other words, the PPD is the marginal distribution of $\vv{v}$ produced by integrating out $\vv{u}$,
\begin{align}
    \pr{\rv{v}|\rv{y}}(\vv{v}|\vv{y})&=
    \int\pr{\rv{x}|\rv{y}}\left(\vv{x}_0(\vv{y})+\mm{W}(\vv{y})\vv{v}+\mm{W}^\perp(\vv{y})\vv{u}\middle|\vv{y}\right)d\vv{u}.
\end{align}

There exist settings in which $f^{\text{sliced}}$ equals $\pr{\rv{v}|\rv{y}}$ (up to a normalization constant). This is the case, for example, when $\rv{u}$ and $\rv{v}$ are statistically independent given $\rv{y}$. However, in imaging inverse problems, this seems to rarely be the case. Indeed, the distribution of natural images is known to be concentrated near a low-dimensional manifold within the high-dimensional ambient space. Therefore, any low-dimensional subspace almost always misses the high-density regions of the distribution. On the other hand, the PPD provides meaningful information regardless of whether the slice passes through the high-density regions of the posterior or not. This is illustrated in Fig.~\ref{fig:proj_vs_slice}.

\begin{figure*}[t]
    \centering
    \includegraphics[width=0.9\linewidth]{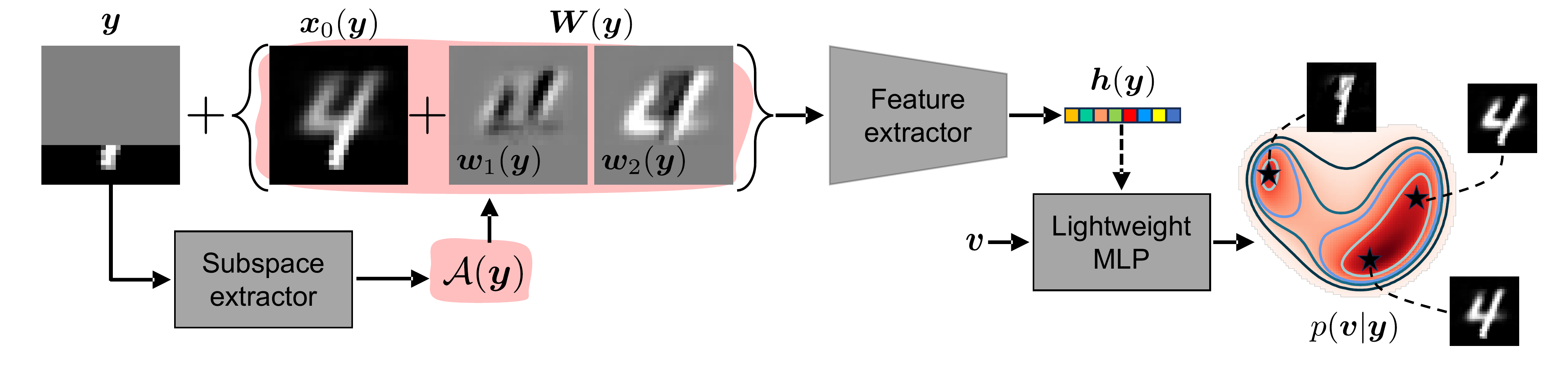}
    \caption{Architecture overview. The degraded image $\vv{y}$ is first fed to a pre-selected subspace extractor that outputs an input-adaptive subspace $\mathcal{A}(\vv{y})=\{\mm{W}(\vv{y}),\vv{x}_0(\vv{y})\}$. Afterward, the degraded image $\vv{y}$ and the extracted subspace are fed to a feature extractor that outputs a feature vector $h(\vv{y})$. The resulting $h(\vv{y})$ modulates a lightweight MLP that outputs $p(\vv{v}|\vv{y})$ for any query $\vv{v}$. The resulting projected distribution can then be navigated to visualize posterior uncertainty.}\label{fig:network}
\end{figure*}

\subsection{Choosing the subspace}\label{subsec:selecting_subspace}

Selecting an appropriate subspace is crucial for producing a meaningful PPD visualization. The ideal subspace should be either 1D or 2D to facilitate visualization. Moreover, the origin $\vv{x}_0(\vv{y})$ and directions $\mm{W}(\vv{y})$ should disclose as much posterior variance as possible, hopefully in a semantically meaningful manner. A natural choice is to take the origin to be the minimum mean square error (MMSE) prediction, $\vv{x}_0(\vv{y})\approx\ee\pbi{\rv{x}|\rv{y}=\vv{y}}$, and the directions $\vv{w}_1(\vv{y}),\dots,\vv{w}_K(\vv{y})$ to be the top $K$ PCs of the posterior. NPPC \cite{nehme2023uncertainty} outputs both $\vv{x}_0(\vv{y})$ and $\vv{w}_1(\vv{y}),\dots,\vv{w}_K(\vv{y})$, rendering it a suitable candidate for fast training and inference. In Fig.~\ref{fig:proj_vs_slice}, we exemplify this choice for the task of image inpainting where moving about the origin $\vv{x}_0(\vv{y})$ along the selected direction $\vv{w}_1(\vv{y})$ changes digit identity.

\subsection{Architecture}\label{subsec:arch}

\begin{figure*}[ht!]
    \centering
    \begin{subfigure}[b]{0.25\linewidth}
    \centering
        \includegraphics[width=0.95\linewidth]{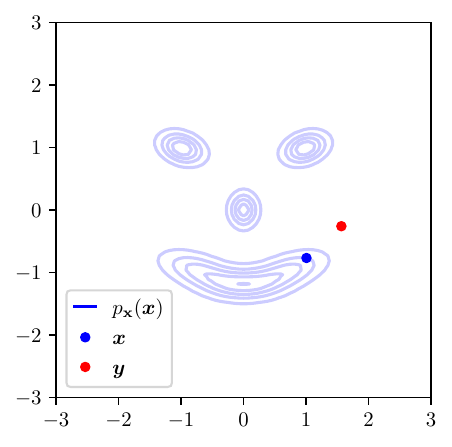}
        \caption{\textcolor{blue}{$\pr{\rv{x}}(\vv{x})$} and a sample $(\textcolor{blue}{\vv{x}_i},\textcolor{red}{\vv{y}_i})$.}\label{fig:2d_model_prior}
    \end{subfigure}
    \begin{subfigure}[b]{0.25\linewidth}
    \centering
        \includegraphics[width=0.95\linewidth]{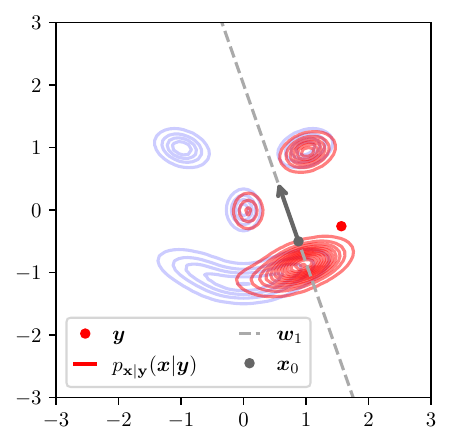}
        \caption{\textcolor{red}{$\pr{\rv{x}|\rv{y}}(\vv{x}|\vv{y})$} and $\mathcal{A}(\vv{y})$.}\label{fig:2d_model_posterior}
    \end{subfigure}
    \begin{subfigure}[b]{0.25\linewidth}
    \centering
        \includegraphics[width=0.95\linewidth]{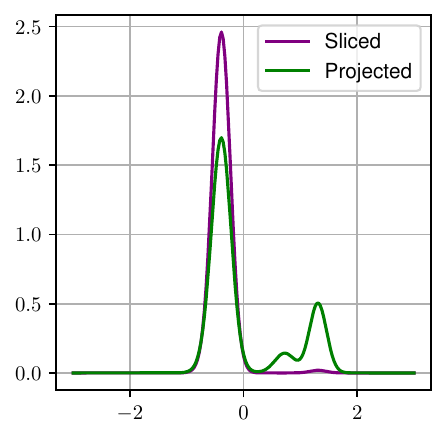}
        \caption{\textcolor{black}{$f^{\text{sliced}}(\vv{v}|\vv{y})$} vs. \textcolor{black}{$\pr{\rv{v}|\rv{y}}(\vv{v}|\vv{y})$}.}\label{fig:2d_model_ppd}
    \end{subfigure}
    \begin{subfigure}[b]{0.25\linewidth}
    \centering
        \includegraphics[width=0.95\linewidth]{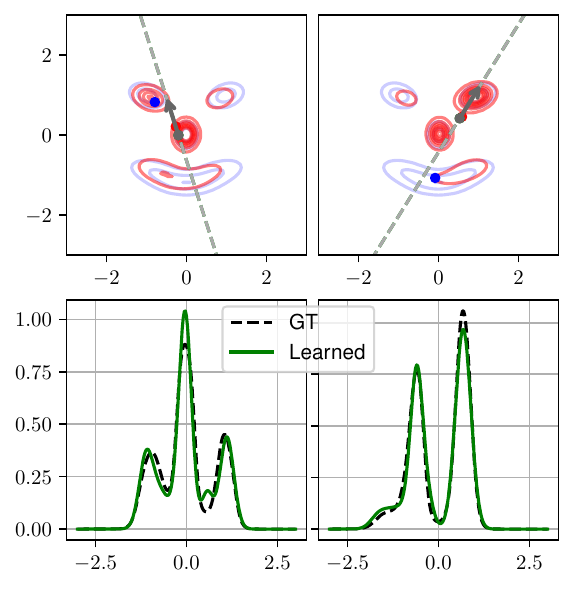}
        \caption{Learned vs. GT $\pr{\rv{v}|\rv{y}}(\vv{v}|\vv{y})$.}
    \end{subfigure}
    \caption{Gaussian mixture denoising. (a) Prior distribution $\pr{\rv{x}}{(\vv{x})}$, and a sample from the joint distribution $\pr{\rv{x},\rv{y}}{(\vv{x},\vv{y})}$. (b) Full posterior distribution $\pr{\rv{x}|\rv{y}}{(\vv{x}|\vv{y})}$, and the selected 1D subspace $\mathcal{A}(\vv{y})=\{\vv{x}_0(\vv{y}),\mm{W}(\vv{y})\}$. (c) Sliced $f^{\text{sliced}}(\vv{v}|\vv{y})$ (purple) vs.~projected $\pr{\rv{v}|\rv{y}}(\vv{v}|\vv{y})$ (green) posterior distribution along the 1D subspace (dashed gray line in (b)). (d) Comparison of the GT and the learned projected posterior.}
    \label{fig:2d_model}
\end{figure*}

We propose to learn the conditional distribution $\pr{\rv{v}|\rv{y}}\ppi{\vv{v}|\vv{y}}$ using a conditional EBM. Unlike normalizing flows, EBMs do not require specialized architectures, and are thus more flexible in their design. EBMs are usually trained with variants of contrastive divergence (CD) \citep{hinton2002training, carreira2005contrastive, bengio2009justifying, sutskever2010convergence} or noise contrastive estimation (NCE) \citep{gutmann2010noise}, with various recent efforts to stabilize and improve training \cite{gao2020learning,du2021improved,zhao2020learning}.

As depicted in Fig.~\ref{fig:network}, we propose to design the architecture using two parts: A (heavy) feature extractor with a classifier-style architecture and a lightweight small MLP with a few linear layers. The feature extractor receives the measurement $\vv{y}$ and the set of vectors defining the subspace $\mathcal{A}(\vv{y})$, and outputs a 1D feature vector $\vv{h}(\vv{y})$. This feature vector is then fed along with a queried point $\vv{v}$ into the second part that outputs $\pr{\rv{v}|\rv{y}}\ppi{\vv{v}|\vv{y}}$. Using this architecture we expect the intermediate feature vector to encapsulate all the information regarding the 2D distribution and have the MLP just translate this data into a distribution function. 

This segregation of the architecture serves the purpose of lowering computational complexity. The rationale behind it is the following: For a given measurement $\vv{y}$, we would like to query $\pr{\rv{v}|\rv{y}}\ppi{\vv{v}|\vv{y}}$ for many values of $\vv{v}$. This is true both during training when running MCMC chains to produce contrastive samples, and during testing when evaluating the PPD on a grid of $\vv{v}$'s for plotting. Therefore, for a given measurement $\vv{y}$ and subspace $\mathcal{A}(\vv{y})$, this inner separation of the architecture allows us to evaluate the feature extractor only once while the lightweight MLP can be queried dozens of times incurring only a minor computational burden. The conditioning mechanism we adopt here is similar to the one employed for timestep encoding in diffusion models \citep{ho2020denoising}, with adaptive shifting and scaling of features reminiscent of AdaIN \citep{huang2017arbitrary}.

To train our conditional EBM, we use a variant of CD \cite{hinton2002training}. In its basic form, CD fits a given parametric function $E\ppi{\vv{v};\vv{\theta}}$ to the (unnormalized) negative log distribution function $-\log \tilde{p}_{\rv{v}}\ppi{\vv{v}}$. In each training step, an MCMC process (\eg Langevin dynamics) is applied to a batch of data samples $\vv{v}$ to produce contrastive samples $\vv{v}_{\text{neg}}$. The model's parameters $\vv{\theta}$ are then updated as
\begin{equation}
\vv{\theta}_{t+1} = \vv{\theta}_t - \nabla_{\vv{\theta}}\left.
\pb{
   \ee_{\vv{v}}\pbi{E\ppi{\vv{v};\vv{\theta}}}-\ee_{\vv{v}_{\text{neg}}}\pbi{E\ppi{\vv{v}_{\text{neg}};\vv{\theta}}}
}\right|_{\vv{\theta}_t}.
\end{equation}
This process gradually increases the likelihood that the model assigns to the samples in the training set and decreases the likelihood that the model assigns to their contrastive counterparts, until convergence. At test time, the resulting model can output the probability density $\pr{\rv{v}}{(\vv{v})}$ (up to an unknown normalization constant) for any queried $\vv{v}$.

The only adjustment needed for using CD to learn a \emph{conditional} distribution, is to add the feature vector $\vv{h}(\vv{y})$ as an additional input to the model, so that $E\ppi{\vv{v},\vv{h}\pp{\vv{y}};\vv{\theta}}=-\log \tilde{p}_{\rv{v}|\rv{y}}\ppi{\vv{v}|\vv{y}}$. Here, we use the CD variant from \cite{yair2022thinking}, which learns a series of distributions by employing multiple levels of additive noise, similarly to diffusion models. 



\section{Experiments}

\begin{figure*}[t!]
    \centering
    \begin{subfigure}[b]{0.33\linewidth}\centering
        \includegraphics[width=0.96\linewidth]{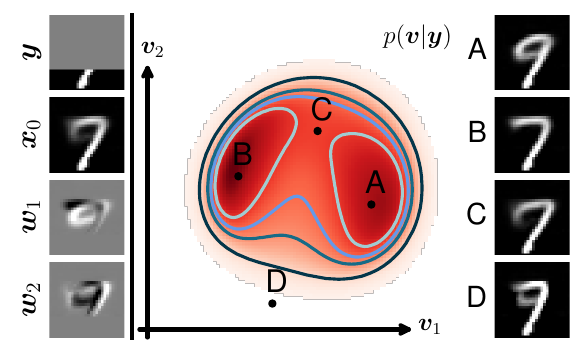}
        \caption{Inpainting}
    \end{subfigure}
    \begin{subfigure}[b]{0.33\linewidth}\centering
        \includegraphics[width=0.96\linewidth]{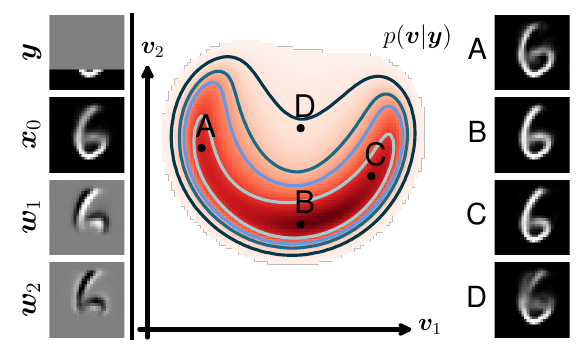}
        \caption{Inpainting}
    \end{subfigure}
    \begin{subfigure}[b]{0.33\linewidth}\centering
        \includegraphics[width=0.96\linewidth]{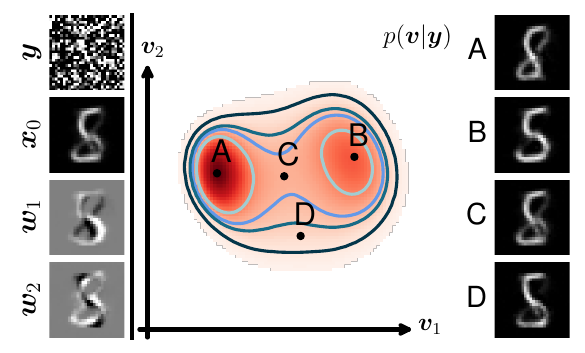}
        \caption{Denoising}
    \end{subfigure}
    \caption{MNIST inpainting and denoising. (a),(b) Application of PPDE to image inpainting from only the 8 bottom pixel rows of the image. The areas within the contours correspond to $50\%,80\%,90\%,98\%$ of the total probability mass. Note how the PPD reveals posterior multi-modality (\eg it shows the digit is either a ``7'' or a ``9''). Similarly, it also reveals intra-digit variations such as the circular part of the digit ``6''. (c) Application of PPDE to image denoising with an extreme noise level of $\sigma_{\varepsilon}=1$. PPDE reveals a similar posterior multi-modality (\eg the digit is either an ``8'' or a ``5'').}
    \label{fig:mnist_results}
\end{figure*}

\begin{figure}
    \centering
    \begin{subfigure}[b]{\linewidth}\centering
        \includegraphics[width=\linewidth]{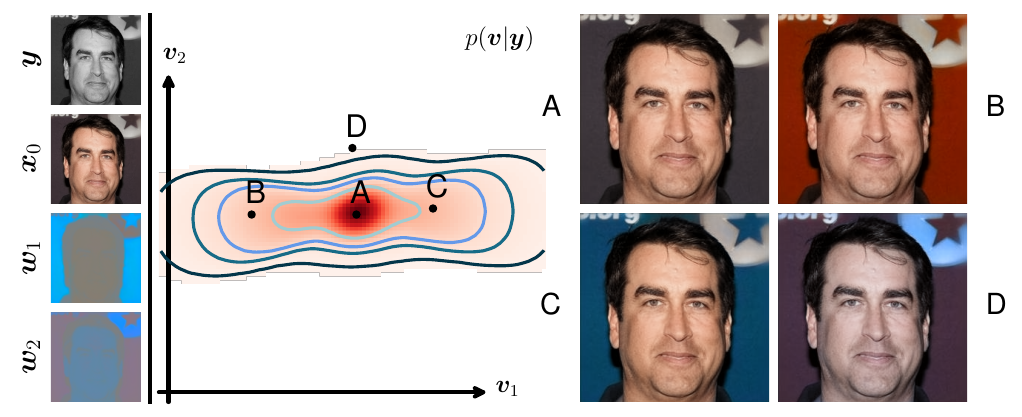}
        \caption{Colorization}
    \end{subfigure}
    \\
    \begin{subfigure}[b]{\linewidth}\centering
        \includegraphics[width=\linewidth]{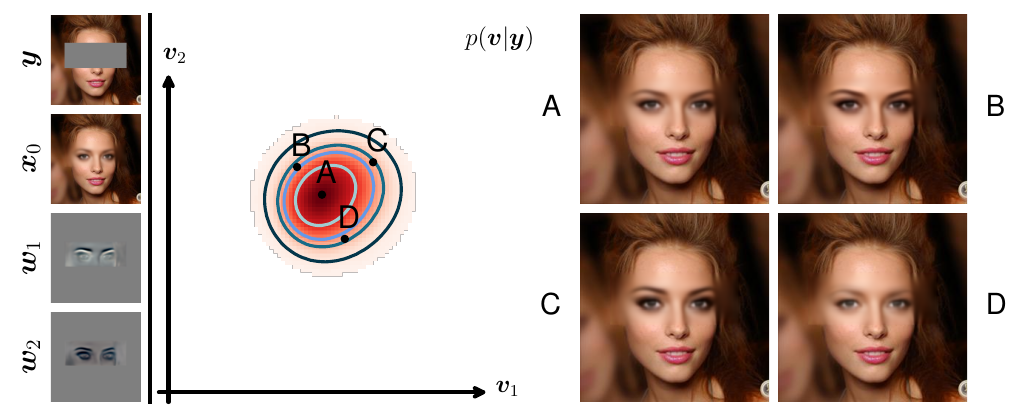}
        \caption{Inpainting}
    \end{subfigure}
    \caption{
        Restoration of face images from CelebA-HQ. The PPD reveals a variety of plausible background and skin colors in colorization (a), and a range of plausible eye-shapes in inpainting (b). In addition, the PPD shows the likelihoods of the different restorations, distinguishing between plausible and less plausible solutions (such as points A and D, respectively, in (a)).
        }  
    \label{fig:celeba_results}
\end{figure}

We now illustrate our method on several tasks and datasets. In all our experiments with visual data, the subspace $\mathcal{A}(\vv{y})$ is spanned by the top two PCs of the posterior, obtained from NPPC \cite{nehme2023uncertainty}. Full details about the architectures, the per-task settings and additional results are in the appendix.


\subsection{Toy example in 2D}

Figure~\ref{fig:2d_model} illustrates the application of PPDE to a 2D toy example with 1D projections. Here, $\rv{x}$ is sampled from a mixture of six Gaussians (arranged as a face), and $\rv{y}$ is a noisy version of $\rv{x}$. The prior distribution $\pr{\rv{x}}(\vv{x})$ and an exemplar sample from the joint distribution $(\vv{x},\vv{y})\sim \pr{\rv{x},\rv{y}}$ are presented in Fig.~\ref{fig:2d_model_prior}. For this simple case, the posterior distribution can be calculated analytically (see appendix for the derivation) and is also a mixture of Gaussians (Fig.~\ref{fig:2d_model_posterior}). Note that for this illustration the posterior itself is two-dimensional, and therefore can be plotted and visualized in full. However, as explained earlier, this is not the case for high-dimensional distributions and is precisely the problem our method aims to solve by projection. Here, this example serves as a sanity check enabling us to benchmark our results against a known ground truth.

To demonstrate our method, we selected an arbitrary 1D subspace on which we project and plot the posterior. The origin point and a direction defining this subspace are shown in Fig.~\ref{fig:2d_model_posterior}. As can be seen in Fig.~\ref{fig:2d_model_ppd}, the PPD on the selected 1D subspace (green line) can provide us with insights regarding the behavior of the full posterior, such as the fact that it is composed of three modes with different weights. In addition, Fig.~\ref{fig:2d_model_ppd} also shows the probability density of the high dimensional (2D) posterior along the 1D line, which we refer to as the sliced posterior (purple line). Comparing both plots, it is easy to see why the sliced posterior is less informative as not all posterior modes appear in the slice projection.

\subsection{Common restoration problems}

\begin{figure}
    \centering
    \includegraphics[width=1.0\linewidth]{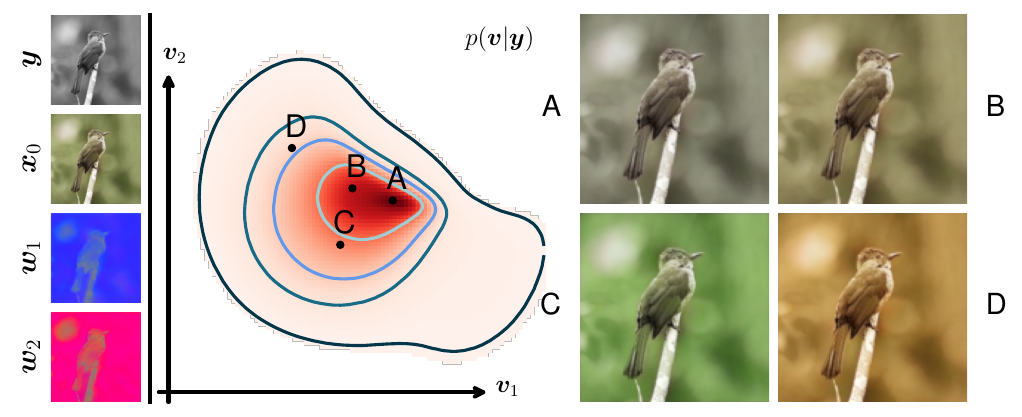} \\
    \includegraphics[width=1.0\linewidth]{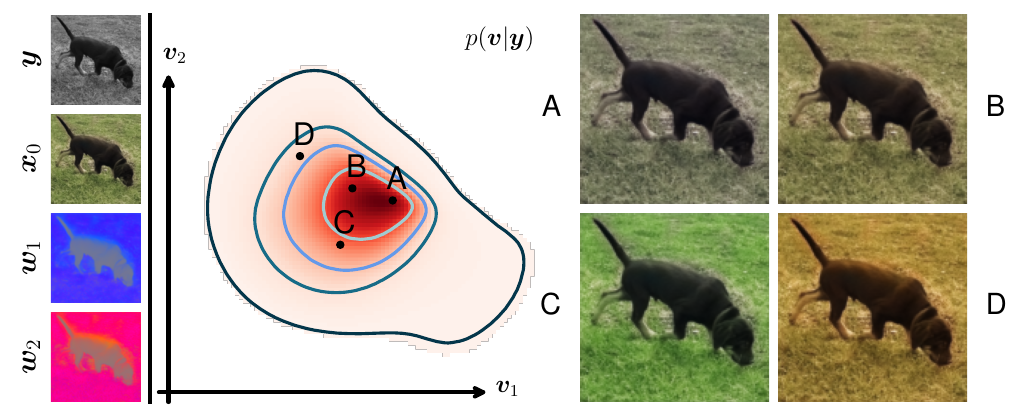}
    \caption{Application of PPDE to natural image colorization on ImageNet, revealing the distribution of different solutions.}
    \label{fig:imagenet-results}
\end{figure}

\begin{figure}
    \centering
    \begin{subfigure}[b]{\linewidth}\centering
        \includegraphics[width=1.0\linewidth]{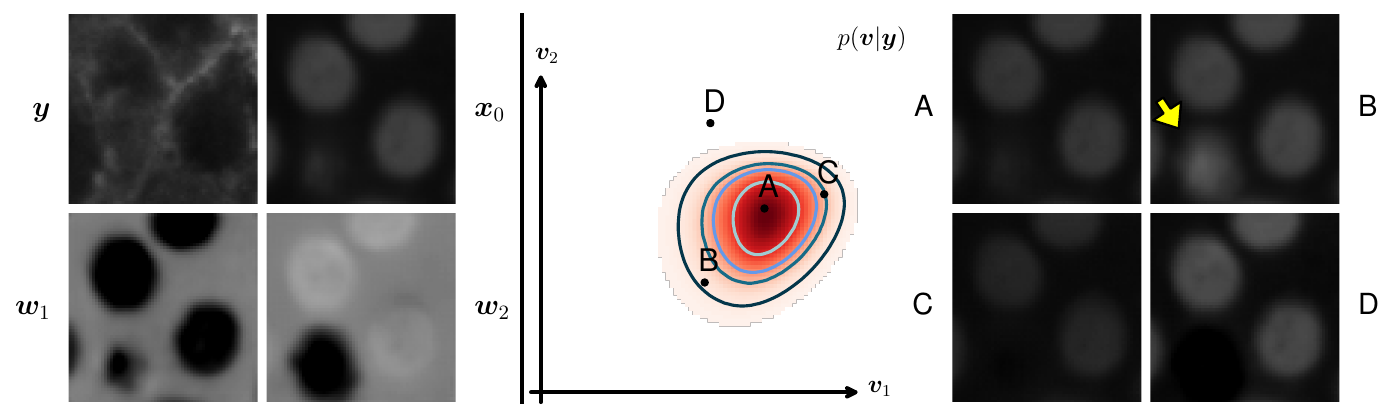}
    \end{subfigure}
    \\
    \begin{subfigure}[b]{\linewidth}\centering
        \includegraphics[width=1.0\linewidth]{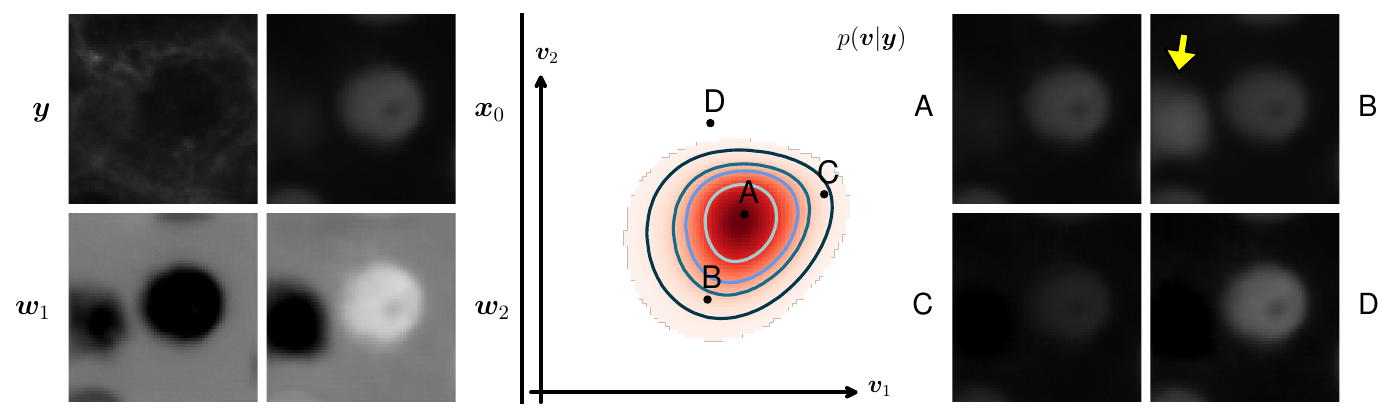}
    \end{subfigure}
    \caption{Biological image-to-image transfer. The PPD outlines the probability of possible solutions for the task of reconstructing fluorescent microscopy images using one type of dye from another. As seen in these examples, the PPD can assist in visualizing the fact that a given input might produce plausible solutions with varying amounts of cells (highlighted by yellow arrows).}
    \label{fig:bio_results}
\end{figure}

\paragraph{Handwritten Digits.}
Figure~\ref{fig:mnist_results} demonstrate PPDE on inpainting and denoising of handwritten digits from the MNIST dataset. In the inpainting task, we used a mask that covers the top $70\%$ of the image, and in denoising we added noise of standard deviation \(\sigma_{\varepsilon}=1\). As can be seen, for both tasks, PPDE reveals the inherent bimodality of the posterior with two different possible digits being likely given the measurement $\vv{y}$. In addition, note that when the digit identity is easier to infer from $\vv{y}$, PPDE reveals uncertainty corresponding to intricate intra-digit variations.

\paragraph{Faces.}
Figure~\ref{fig:celeba_results} presents results for colorization and inpainting of face images from the CelebA-HQ \(256\times 256\) dataset. In the appendix, we further provide results for $8\times$ super-resolution with a bicubic downsampling filter and a crop area taken from SRFlow \citep{lugmayr2020srflow}. As can be seen, PPDE reveals the unique structure of the projected posterior distribution, which outlines the region of plausible solutions in the projected space.


\paragraph{Natural images.}
Figure~\ref{fig:imagenet-results} shows colorization results for natural images from the ImageNet 1K dataset \citep{russakovsky2015imagenet}. The distributions in this case are often not multi-modal; however, they are also far from being axis-aligned Gaussians. This demonstrate the practical benefit of PPD visualizations. More examples are available in the Appendix.

\begin{figure*}
    \centering
    \includegraphics[width=0.9\linewidth]{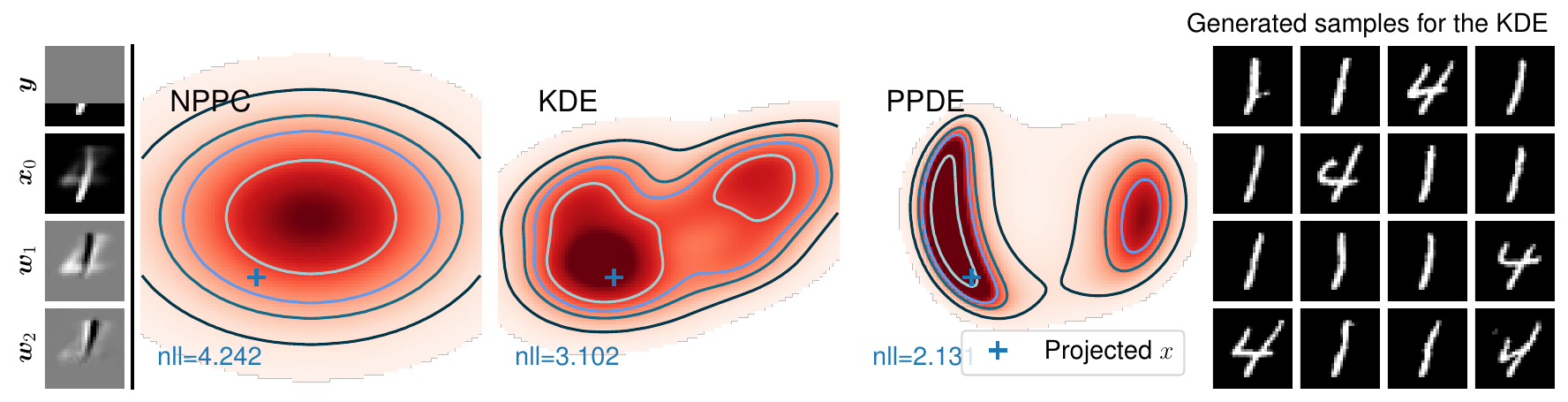}
    \caption{PPD comparison on MNIST. On the left we plot the input measurement $\vv{y}$, and the input-adaptive subspace $\mathcal{A}(\vv{y})=\bri{\vv{x}_0(\vv{y}),\vv{w}_1(\vv{y}),\vv{w}_2(\vv{y})}$. In the middle, we plot the estimated PPDs obtained from a Gaussian approximation using NPPC, from the KDE approach, and from our PPDE. The projected ground truth posterior sample $\vv{x}$ is marked by a blue cross, and its likelihood under the estimated PPD is shown on the bottom left of the corresponding plot. On the right, we show a few posterior samples out of the 100 used in the KDE approach. These were obtained using an EBM trained on MNIST.}
    \label{fig:ppd-quantitative-mnist}
\end{figure*}

\begin{figure*}
    \centering
    \includegraphics[width=0.9\linewidth]{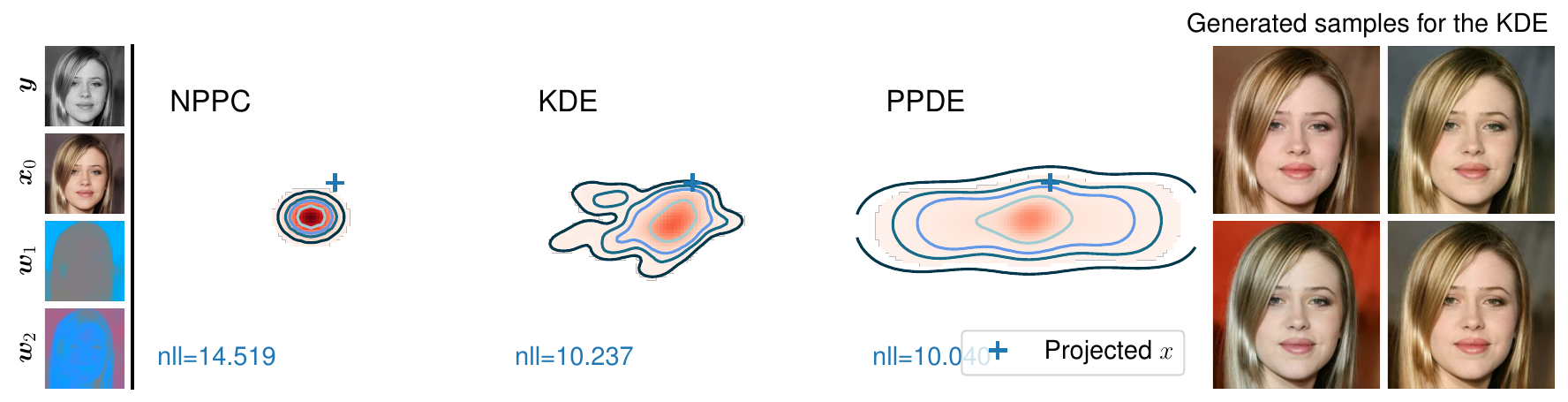}
    \caption{PPD comparisons on CelebA-HQ. The layout is the same as Fig.~\ref{fig:ppd-quantitative-mnist}. The samples on the right were obtained using DDNM.}
    \label{fig:ppd-quantitative-celebahq}
\end{figure*}

\paragraph{Biological image-to-image translation.}
Finally, we test our method on a nonlinear image-to-image translation task. Here, we evaluate PPDE on the dataset from~\citep{von2021democratising}, where a microscopic biological specimen imaged with one fluorescent dye is transformed to appear as if it was imaged with another. This ``virtual staining'' task is of immense importance in biological imaging, as it enables correlative imaging using fewer fluorescent dyes. Figure~\ref{fig:bio_results} presents the results of PPDE on this challenging task. The resulting distribution discloses a range of plausible solutions in the projected space corresponding to different numbers of cells. This example demonstrates how PPDE can be used as a quantitative tool for assessing the variance in cell counting applications.

\subsection{Quantitative comparisons}

\paragraph{Comparison to NPPC.}
NPPC outputs the posterior mean $\vv{x}_0(\vv{y})$, alongside the top $K$ posterior PCs $\mm{W}(\vv{y})$ and variances $\{\sigma_k^2(\vv{y})\}_{k=1}^K$ for a given input image $\vv{y}$. While it was originally proposed as a distribution-free method, the predicted variances can be used to construct a 2D Gaussian approximation of the projected posterior covariance. Let $\mm{\Sigma}(\vv{y})$ denote the diagonal matrix that has $[\sigma_1^2(\vv{y}),\sigma_2^2(\vv{y})]^{\top}$ along its main diagonal. Then, the projected posterior distribution can be approximated as
\begin{equation}
    \pr{\rv{v}|\rv{y}}(\vv{v}|\vv{y})\approx\mathcal{N}(\vv{v};0,\mm{\Sigma}(\vv{y})).
\end{equation}
We compare our results to this baseline in Table~\ref{table:results}, showing a significant improvement in sample negative log likelihood (NLL) in favor of PPDE.

\begin{table}
\centering
\caption{Negative log likelihood comparison. NPPC and our method were computed on the entire test set. The KDE approach was computed for 100 test images due to computational constraints.}
\setlength{\tabcolsep}{2pt}
\label{table:results}
\begin{tabular}{l c c c} 
    \hline\\
    Task & NPPC & KDE & Ours\\
    & (full test) & (100 samples) & (full test) \\
    
    \hline \\
    {\bf MNIST} &&  \\ 
    \hline
    Inpainting & 4.19 ± 0.02 & 3.9 ± 0.09 & \textbf{3.72 ± 0.01} \\
    Denoising & 3.41 ± 0.02 & 3.26 ± 0.07 & \textbf{3.04 ± 0.01} \\

    \hline \\
    {\bf CelebA} &&  \\
    \hline
    Inpainting Eyes &  6.38 ± 0.08 & 8.278 ± 1.1 & \textbf{5.37 ± 0.03} \\
    Colorization & 10.3 ± 0.2 & 7.7 ± 0.3 & \textbf{7.39 ± 0.03} \\
    Super-resolution & 5.12 ± 0.3 & - & \textbf{2.84 ± 0.1} \\


    \hline \\
    {\bf ImageNet} &&  \\
    \hline
    Colorization & 10.15 ± 0.06 & - & \textbf{6.86 ± 0.01} \\
    \hline
\end{tabular}
\end{table}

\paragraph{Comparison to posterior sampling + KDE.}
The aim of our density estimator is to estimate the PPD in one or two dimensions. Given a method that can sample from the posterior (\eg DDNM \citep{wang2023ddnm}), a possible baseline approach to achieve this goal is as follows: (i) Generate $N$ samples from the posterior $\{\vv{x}_i\}$; (ii) Compute PCA using $\{\vv{x}_i\}$ and project the samples onto the space spanned by the top two PCs; (iii) Estimate the PPD in 2D from the projected samples $\{\vv{v}_i\}$ using standard techniques such as \emph{kernel density estimation} (KDE). Comparisons against this baseline are reported in Table~\ref{table:results}. For producing the KDE samples from the posterior we have used: a conditional EBM for MNIST, DPS \citep{chung2022diffusion}\footnote{For DPS we have used the public model trained on FFHQ (and not on CelebA). Therefore, the model is not expected to produce accurate samples. The comparison to the PPD it produces is presented here only as an additional rough baseline.}
for face inpainting\footnote{We also experimented with MAT \citep{li2022mat}, which gave inferior results.}, and DDNM for face colorization, selecting the best sampler according to their performance and available trained models. For each task/dataset we randomly chose 100 test images and sampled 100 posterior samples per image. Compared to this (impractical) baseline, our technique is orders of magnitude faster as this approach requires costly repeated posterior sampling to estimate the projected density with enough accuracy. Figures~\ref{fig:ppd-quantitative-mnist} and ~\ref{fig:ppd-quantitative-celebahq} present visual examples of the approximated PPD with NPPC, KDE, and PPDE. PPDE exhibits superior NLL on average while being extremely faster than KDE, and only slightly slower than NPPC.

\section{Discussion and conclusion}

We presented a new method for \emph{informed} uncertainty visualization and demonstrated its practical benefit across various tasks and datasets. Compared to the proposed baselines, our method brought a significant advantage both qualitatively and quantitatively in terms of improved sample NLL. The main limitation of our method lies in the fact that visualization beyond two or three dimensions becomes very difficult. This limits our supported analysis to 2 or 3 PCs, which may be insufficient to faithfully represent rich posteriors. Our method is particularly suited for tasks where posteriors have a spectrally-concentrated covariance, such as in image colorization. However, for severely ill-posed inverse problems where a large number of PCs is required to project the posterior in an informative manner, our strategy becomes impractical, warranting further research of condensed alternative representations of uncertainty.

\section*{Acknowledgements}

This research was partially supported by the Israel Science Foundation (grant no. 2318/22), by a gift from Elbit Systems, and by the Ollendorff Minerva Center, ECE faculty, Technion.

\clearpage
\appendix

\section{Toy model}

\subsection{Problem setting} \label{app:prob-setting-2d}

In the main text, we demonstrated PPDE on a 2D denoising task, where samples $\vv{x}_i$ from a Gaussian mixture model were distorted by additive white Gaussian noise $\vv{n}_i$ with a standard deviation of $\sigma_n$ to result in noisy measurements $\vv{y}_i$. The prior distribution $\pr{\rv{x}}{(\vv{x})}$ was comprised of $L$ Gaussians defined by the parameters $\br{\vv{\mu}_\ell,\mm{\Sigma}_\ell,\pi_\ell}_{\ell=1}^L$, namely,
\begin{equation}
\pr{\rv{x}}\ppi{\vv{x}}
=\sum_{\ell=1}^L \pi_\ell\cdot\varphi\ppi{\vv{x};\vv{\mu}_\ell,\mm{\Sigma}_\ell},
\end{equation}
where $\varphi\ppi{\vv{x};\vv{\mu},\mm{\Sigma}}$ is the PDF of a multivariate normal distribution with mean $\vv{\mu}$ and covariance matrix $\mm{\Sigma}$. Specifically, in the 2D example of Fig.~5 from the main text, we used $L=6$ Gaussians with the following parameters:
\begin{equation}
    \begin{array}{lll}
        \vv{\mu}_1 = \ppi{-1, 1}^T &
        \Sigma_1 = \begin{pmatrix} 0.05 & -0.01 \\ -0.01 & 0.025 \end{pmatrix} &
        \pi_1 = 0.15
        \\
        \vv{\mu}_2 = \ppi{1, 1}^T &
        \Sigma_2 = \begin{pmatrix} 0.05 & 0.01 \\ 0.01 & 0.025 \end{pmatrix} &
        \pi_2 = 0.15
        \\
        \vv{\mu}_3 = \ppi{0, 0}^T &
        \Sigma_3 = \begin{pmatrix} 0.02 & 0 \\ 0 & 0.03 \end{pmatrix} &
        \pi_3 = 0.15
        \\
        \vv{\mu}_4 = \ppi{-0.7, -1}^T &
        \Sigma_4 = \begin{pmatrix} 0.15 & -0.04 \\ -0.04 & 0.04 \end{pmatrix} &
        \pi_4 = 0.15
        \\
        \vv{\mu}_5 = \ppi{0, -1.2}^T &
        \Sigma_5 = \begin{pmatrix} 0.15 & 0 \\ 0 & 0.025 \end{pmatrix} &
        \pi_5 = 0.15
        \\
        \vv{\mu}_6 = \ppi{0.7, -1}^T &
        \Sigma_6 = \begin{pmatrix} 0.15 & 0.04 \\ 0.04 & 0.04 \end{pmatrix} &
        \pi_6 = 0.15.
        \\
    \end{array}
\end{equation}
The standard deviation of the additive noise was $\sigma_n=0.4$.

\subsection{Analytical posterior distribution}

Assuming the setting in Sec.~\ref{app:prob-setting-2d}, here we derive the analytical posterior distribution. Specifically, we show that the posterior is also a GMM, and that it has the form
\begin{equation}
\pr{\rv{x}|\rv{y}}\ppi{\vv{x}|\vv{y}}
=\sum_{\ell=1}^L \tilde{\pi}_\ell\ppi{\vv{y}}\cdot\varphi\ppi{\vv{x};\tilde{\vv{\mu}}_\ell\ppi{\vv{y}},\tilde{\mm{\Sigma}}_\ell(\vv{y})},\label{eq:gmm-posterior}
\end{equation}
where
\begin{align}
\tilde{\vv{\mu}}_\ell\pp{\vv{y}}&=\vv{\mu}_\ell+\mm{\Sigma}_\ell\ppi{\mm{\Sigma}_\ell + \sigma_n^2\mm{I}}^{-1}\pp{\vv{y}-\vv{\mu}_\ell}, \nonumber\\ 
\tilde{\mm{\Sigma}}_\ell(\vv{y})&=\mm{\Sigma}_\ell-\mm{\Sigma}_\ell\ppi{\mm{\Sigma}_\ell + \sigma_n^2\mm{I}}^{-1}\mm{\Sigma}_\ell, \label{eq:tilde_mu_sigma}\nonumber\\
\tilde{\pi}_\ell\ppi{\vv{y}}
&=\frac
{\pi_\ell \varphi\ppi{\vv{y};\vv{\mu}_\ell,\mm{\Sigma}_\ell+\sigma_n^2\mm{I}}}
{\sum_{\ell'=1}^L\pi_{\ell'} \varphi\ppi{\vv{y};\vv{\mu}_{\ell'},\mm{\Sigma}_{\ell'}+\sigma_n^2\mm{I}}}. 
\end{align}

To derive this result, we start by invoking the law of total probability, marginalizing over an auxiliary random variable $\rn{c}$ that selects one of the $L$ distributions with probabilities $\br{\pi_1,\dots,\pi_L}$,
\begin{equation}
\pr{\rv{x}}\ppi{\vv{x}}
=\sum_{\ell=1}^L \pr{\rv{x}|\rn{c}}\ppi{\vv{x}|\ell}\pr{\rn{c}}\ppi{\ell}.
\end{equation}
The posterior distribution can then be written as
\begin{equation}
\pr{\rv{x}|\rv{y}}\ppi{\vv{x}|\vv{y}} = \sum_{\ell=1}^L \pr{\rv{x}|\rv{y},\rn{c}}\ppi{\vv{x}|\vv{y},\ell}\pr{\rn{c}|\rv{y}}\ppi{\ell|\vv{y}}. \label{eq:post_dist}
\end{equation}
The term $\pr{\rn{c}|\rv{y}}\ppi{\ell|\vv{y}}$ in Eq.~\eqref{eq:post_dist} can be written using Bayes' rule as
\begin{align}
\tilde{\pi}_\ell\ppi{\vv{y}}&\triangleq
\pr{\rn{c}|\rv{y}}\ppi{\ell|\vv{y}}\nonumber\\
&=\frac
{\pr{\rv{y}|\rn{c}}\ppi{y|\ell}\pr{\rn{c}}\ppi{\ell}}
{\pr{\rv{y}}\ppi{\vv{y}}} \nonumber\\
&=\frac
{\pr{\rv{y}|\rn{c}}\ppi{\vv{y}|\ell}\pr{\rn{c}}\ppi{\ell}}
{\sum_{\ell'=1}^L\pr{\rv{y}|\rn{c}}\ppi{\vv{y}|\ell'}\pr{\rn{c}}\ppi{\ell'}} \nonumber\\
&=\frac
{\pi_\ell \varphi\ppi{\vv{y};\vv{\mu}_\ell,\mm{\Sigma}_\ell+\sigma_n^2\mm{I}}}
{\sum_{\ell'=1}^L\pi_{\ell'} \varphi\ppi{\vv{y};\vv{\mu}_{\ell'},\mm{\Sigma}_{\ell'}+\sigma_n^2\mm{I}}}.
\end{align}
Moreover, to explicitly express $\pr{\rv{x}|\rv{y},\rn{c}}\ppi{\vv{x}|\vv{y},\ell}$ in Eq.~\eqref{eq:post_dist}, we note that $\rv{x}$ and $\rv{y}$ are jointly Gaussian given $\rn{c}$, because under a particular choice $\rn{c}=\ell$, the random vector $\rn{y}$ is the sum of two independent Gaussian vectors, $\rv{x}$ and $\rv{n}$. Hence, conditioned on the event $\rn{c}=\ell$, the joint distribution of $\rv{x}$ and $\rv{y}$ is given by
\begin{align}
\pr{\rv{x},\rv{y}|\rn{c}}\ppi{\vv{x},\vv{y}|\ell}&=\mathcal{N}\pp{
    \begin{pmatrix}\vv{\mu}_\ell \\ \vv{\mu}_\ell\end{pmatrix},
    \begin{pmatrix}
        \mm{\Sigma}_{\rv{x}\rv{x}| \rn{c}}(\ell) & \mm{\Sigma}_{\rv{x}\rv{y}| \rn{c}}(\ell) \\
        \mm{\Sigma}_{\rv{x}\rv{y}| \rn{c}}(\ell) & \mm{\Sigma}_{\rv{y}\rv{y}| \rn{c}}(\ell)
    \end{pmatrix}
} \nonumber\\
& = \mathcal{N}\pp{
    \begin{pmatrix}\vv{\mu}_\ell \\ \vv{\mu}_\ell\end{pmatrix},
    \begin{pmatrix}
        \mm{\Sigma}_\ell & \mm{\Sigma}_\ell \\
        \mm{\Sigma}_\ell & \mm{\Sigma}_\ell + \sigma_n^2\mm{I}
    \end{pmatrix}
}.\label{eq:gmm-xy-joint}
\end{align}
From the joint distribution in Eq.~\eqref{eq:gmm-xy-joint}, the conditional distribution of $\rv{x}$ given $\rv{y}$ and $\rn{c}$ is derived as
\begin{equation}
\pr{\rv{x}|\rv{y},\rn{c}}\ppi{\vv{x}|\vv{y},\ell}
=\varphi\ppi{\vv{x};\tilde{\vv{\mu}}_\ell\ppi{\vv{y}},\tilde{\mm{\Sigma}}_\ell(\vv{y})},
\end{equation}
with $\tilde{\vv{\mu}}_\ell\ppi{\vv{y}}$ and $\tilde{\mm{\Sigma}}_\ell(\vv{y})$ defined as in Eq.~\eqref{eq:tilde_mu_sigma}.

\subsection{Projected posterior distribution}
Given that the posterior distribution is a GMM as in Eq.~\eqref{eq:gmm-posterior}, it is straightforward to derive the \emph{projected posterior distribution} (PPD) over any arbitrary 1D subspace defined by a center $\vv{x}_0$ and a normalized direction $\vv{w}$. This is done by projecting each of the Gaussians individually, \ie the PPD for $v=\vv{w}^T\ppi{\vv{x}-\vv{x}_0}$ is given by
\begin{equation}
\pr{\rn{v}|\rv{y}}\ppi{v|\vv{y}}
=\sum_{\ell=1}^L \tilde{\pi}_\ell\ppi{\vv{y}}\cdot
\varphi\ppi{v;
\vv{w}^T\ppi{\tilde{\vv{\mu}}_\ell\ppi{\vv{y}}-\vv{x}_0}
,\vv{w}^T\tilde{\mm{\Sigma}}_\ell(\vv{y})\vv{w}}.
\end{equation}

During both training and testing, we took $\vv{x}_0$ to be the posterior mean for a given $\vv{y}$,
\begin{equation}
\vv{x}_0(\vv{y})=\sum_{\ell=1}^L \tilde{\vv{\mu}}_\ell\ppi{\vv{y}},
\end{equation}
and $\vv{w}$ to be a random normalized direction.

\section{PPD learning with an EBM}

For the purpose of training the conditional EBM, we used a variant of contrastive divergence (CD), proposed in \citep{yair2022thinking}. This method attempts to model the log distribution of the dataset by employing a series of distributions that gradually transition between the distribution of the dataset to some reference Gaussian distribution with known parameters. We refer to this method here as \emph{multilevel CD} (MCD) and review it only briefly for completeness. For a more detailed description please refer to \citep{yair2022thinking}.

At the core of MCD lies a series of distributions defined by (i) coefficients $\br{\alpha_t}_{t=1}^T$ gradually decreasing from $\alpha_1=1$ to $\alpha_T=0$, (ii) a discrete random variable $\rn{t}\in\br{1,\dots,T}$, with some chosen prior distribution $\pr{\rn{t}}$, and (iii)~a reference Gaussian vector $\rv{n}$. During training, a new random variable $\tilde{\rv{x}}$ is defined as a linear mixture of $\rv{x}$ and $\rv{n}$ with random linear coefficients associated with $\alpha_\rv{t}$ (note that $\rv{t}$ is a random variable),
\begin{equation}
\tilde{\rv{x}}=\alpha_\rv{t} \rv{x} + \sqrt{1 - \alpha_\rv{t}^2}\, \rv{n}.
\end{equation}
The surrogate objective of MCD is to model the mixed distribution of the pair $\ppi{\tilde{\rv{x}},\rn{t}}$. This is done using a neural network $\vv{f}\ppi{\tilde{\vv{x}};\vv{\theta}}$ with parameters $\vv{\theta}$, which receives $\tilde{\vv{x}}$ as input, and outputs $T$ scalars, where the $t$-th output of the network models the log-probability
\begin{equation}
f_t\ppi{\tilde{\vv{x}};\vv{\theta}}
\approx\log\pr{\tilde{\rv{x}},\rn{t}}\ppi{\tilde{\vv{x}},t}
=\log\pr{\tilde{\rv{x}}|\rn{t}}\ppi{\tilde{\vv{x}}|t}+\log\pr{\rn{t}}\ppi{t}.
\end{equation}
The distribution of $\rv{x}$ can then be extracted from the model at $t=1$ as
\begin{equation}
f_1\ppi{\vv{x};\vv{\theta}}
\approx\log\pr{\rv{x}}\ppi{\vv{x}}+\log\pr{\rn{t}}\ppi{1}.
\end{equation}

The motivation for modeling the pair $\ppi{\tilde{\rv{x}},\rn{t}}$, as opposed to just modeling $\rv{x}$, is to improve the accuracy of CD in modeling low-density regions of $\log\pr{\rv{x}}{(\vv{x})}$. During training, the model is only exposed to samples $\vv{x}$ coming from regions of high probability. Therefore, without further modifications, it usually fails to extrapolate the density function beyond these regions correctly. Training on pairs $\ppi{\tilde{\rv{x}},\rn{t}}$ generated by sampling $\vv{n}$ and $t$ from their known distribution and $\vv{x}$ from the dataset, MCD can better overcome this challenge. 

The update step of MCD combines the standard CD update at any fixed $t$ with a classification update for $t$ given $\tilde{\vv{x}}$. The CD update at a fixed $t$ is derived by applying the classical CD algorithm to the $t$-th output of the network, which models $\log\pr{\tilde{\rv{x}}|\rn{t}}$ (up to the known term $\log\pr{\rn{t}}\ppi{t}$). At each training step, the modeled $\log\pr{\tilde{\rv{x}}|\rn{t}}$ is used to produce an MCMC process starting at $\tilde{\vv{x}}$ and ending at a contrastive sample $\tilde{\vv{x}}_{\text{neg}}$. The used MCMC process was Langevin dynamics \cite{welling2011bayesian} with a Metropolis-Hastings rejection step similar to \cite{metropolis1953equation,hastings1970monte}. The parameters of the model were then updated according to the CD algorithm, following the negative gradient of $f_t\ppi{\tilde{\vv{x}}_{\text{neg}};\vv{\theta}}-f_t\ppi{\tilde{\vv{x}};\vv{\theta}}$.

As for the classification update, it was derived by viewing the network as a classifier, minimizing the cross-entropy loss. This holds since a $\softmax$ operation on the network outputs results in the estimation of the conditional distribution $\pr{\rn{t}|\tilde{\rv{x}}}$
\begin{equation}
\softmax\ppi{\vv{f}\ppi{\tilde{\vv{x}};\vv{\theta}}}_t
\approx\frac
{\pr{\tilde{\rv{x}},\rn{t}}\ppi{\tilde{\vv{x}},t}}
{\sum_{t'=1}^T \pr{\tilde{\rv{x}},\rn{t}}\ppi{\tilde{\vv{x}},t'}}=\pr{\rn{t}|\tilde{\rv{x}}}\ppi{t|\tilde{\vv{x}}}.
\end{equation}




The combined update rule of MCD was therefore given by
\begin{align}
\vv{\theta}^{\ppi{k+1}}
=\vv{\theta}^{\ppi{k}}-\eta\nabla_{\vv{\theta}}
&\left[
f_t\ppi{\tilde{\vv{x}}_{\text{neg}};\vv{\theta}}-f_t\ppi{\tilde{\vv{x}};\vv{\theta}}
\right.
\nonumber \\
&\qquad\left.
+\softmax\ppi{\vv{f}\ppi{\tilde{\vv{x}};\vv{\theta}}}_t
\right],
\end{align}
where $\eta$ is the algorithm's learning rate.

Note that thus far the derivation of MCD assumed an unconditional setting where our goal was to model $\log\pr{\rv{x}}\ppi{\vv{x}}$. To adopt this training scheme for modeling the PPD, we need to replace $\vv{x}$ with its projection $\vv{v}=\mm{W}^T\ppi{\vv{x}-\vv{x}_0}$ and make the EBM depend on the distorted sample $\vv{y}$ and the selected subspace $\mathcal{A}\ppi{\vv{y}}=\br{\vv{x}_0\ppi{\vv{y}},\mm{W}\ppi{\vv{y}}}$. As described in Sec.~3.3 of the main text, this dependency is achieved using a separate feature extractor which outputs a feature vector $\vv{h}$ conditioning the layers of the EBM through feature normalization.


\section{Datasets and restoration tasks}

We evaluated our method on a range of datasets and tasks spanning common scenarios in low-level vision. Specifically, we used the MNIST dataset \citep{deng2012mnist} ($28\times 28$), the CelebA dataset \citep{liu2015faceattributes} (cropped and resized to $160\times 160$ as in SRFlow \citep{lugmayr2020srflow}), the CelebA-HQ dataset \citep{karras2017progressive} ($256\times 256$), the ImageNet 1K dataset \cite{russakovsky2015imagenet} ($128\times 128$) and an a cells microscopy dataset presented in \cite{von2021democratising}. The tasks we experimented with were:

\begin{itemize}
    \item \textbf{MNIST inpainting}. Recovering digit images from only the bottom 8 rows.
    \item \textbf{MNIST denoising}. Denoising digit images contaminated with additive white Gaussian noise with $\sigma_n=1$ and clipping.
    \item \textbf{CelebA-HQ inpainting}. Recovering a rectangular area of size $70\times175$ around the eyes in CelebA-HQ images of size $256\times 256$.
    \item \textbf{CelebA-HQ colorization}. Recovering RGB images from grayscale images obtained by averaging color channels.
    \item \textbf{CelebA super-resolution}. 8$\times$ super-resolution of face images downsampled with a bicubic kernel to $20 \times 20$ pixels as described in \citep{lugmayr2020srflow}.
    \item \textbf{ImageNet colorization}. Recovering RGB images from grayscale images obtained by averaging color channels.
    \item \textbf{Biological image-to-image}. Transforming a given microscopic biological imaged specimen with one fluorescent dye appears as if it was imaged by another. We have broken this task into transforming patches of size $64\times64$ between the 2 domains. 
\end{itemize}


\section{Architectures and training hyperparameters}

The estimation of the projected posterior distribution (PPD) requires the selection of a subspace for each distorted image on which the posterior is to be projected. As described in Sec.~3.2 of the main text, we selected in this work to use a subspace that contains the minimum MSE (MMSE) estimator and is spanned by the first two principal components (PCs) of the posterior. For the purpose of predicting these values we trained two auxiliary networks. Overall, we have sequentially trained three networks for each of the evaluated tasks.

The first network was trained using an MSE loss to output the MMSE estimate for a given distorted image $\vv{y}$. The second network was trained using NPPC \cite{nehme2023uncertainty} to predict the first five principal components of the posterior given both the distorted image and the MMSE estimate (predicted by the first network). The third network was trained using the method described in this paper to model the PPD.

Each network required a different architecture depending on the task at hand. Overall, for every task, we used a specific subset of the following architectures:
\begin{itemize}
    \item \textbf{U-Net-S}. A small version of the original U-Net architecture from \cite{ronneberger2015u} with three encoder and decoder levels containing a single convolution layer each. The number of channels in each level is $32$, $64$ and $128$, respectively. The bottleneck has two convolutions of $256$ channels. In addition, we used a group-norm layer after each convolution, LeakyReLU as a non-linearity layer, and a nearest-neighbor interpolation for upsampling.

    \item \textbf{U-Net-L}. A U-Net architecture similar to the one used in DDPM \cite{ho2020denoising} for the CelebA generative task, but with half the number of channels. We also removed all network parts related to the time index dependency.

    \item \textbf{U-Net-M}.
    A network similar to U-net-L with one level less in the encoder and decoder (the top level, in terms of the number of channels).

    \item \textbf{EDSR}.    
    An architecture similar to the EDSR architecture described in \cite{lim2017enhanced} using 16 residual blocks and a width of 64 channels.

    \item \textbf{cEBM-S}.    
    This network is comprised of two parts, as described in Sec.~3.3 in the main text. The first part is a feature extractor based on a ResNet18 architecture \cite{he2016deep}, fitted for smaller images. We removed the stride in the input convolution and the following pooling layer and reduced the number of channels by 2$\times$. In addition, we replaced the BatchNorm layers with GroupNorm layers and performed the downsampling using average pooling layers (instead of stride convolutions). The output of this network is a feature vector of length 256. The second part of the network is a conditional MLP with 6 hidden layers of width 128. We used SiLU as the non-linearity followed by an AdaIN \cite{huang2017arbitrary} layer that performs adaptive scaling and shifting. The scaling and shifting parameters are extracted from the feature vector using a linear layer.
    
    \item \textbf{cEBM-L}. This network is similar to the cEBM-S except that the architecture of the feature extractor is replaced by the encoder part of the U-Net-L, using a global average pooling layer at the end of the bottleneck to produce a feature vector of length 512.
    
\end{itemize}

In general, the U-Net architectures were used for the MMSE estimator and the PCs, and the conditional EBMs was used for modeling the PPD. The only exceptions were the networks in the super-resolution task, in which the distorted image was smaller than all other images and therefore was fed into the EDSR architecture either to directly produce the desired output (as in the case of the MMSE) or as a preprocessing step before concatenating it with the other inputs.

All networks were trained using the Adam optimizer, stopping the training process once it reached a minimal value of the objective over a dedicated validation set. For the training of the PC predictor and the PPD model, the learning rate was kept fixed. On the other hand, for training the MMSE estimator, the learning rate was reduced by half every 5000 steps. Table \ref{table:architertures} summarizes the architectures and training hyperparameters used for the networks in each task/dataset.

\begin{table*}
\centering
\caption{Architectures, learning rates, and number of steps used per task/dataset.}
\setlength{\tabcolsep}{2pt}
\label{table:architertures}
\begin{tabular}{l | c c c | c c c | c c c} 
    \hline &&&&&&&&&\\
    & MMSE & MMSE & MMSE & PCs & PCs & PCs & PPD & PPD & PPD \\
    Task & Arch. & LR & \# of steps & Arch. & LR & \# of steps & Arch. & LR & \# of steps \\
    
    \hline &&&&&&&&& \\
    {\bf MNIST} &&&&&&&&& \\ 
    \hline
    Inpainting
    & U-Net-S & $10^{-4}$ & 3,510
    & U-Net-S & $10^{-4}$ & 5,000
    & cEBM-S & $10^{-4}$ & 36,504
    \\
    Denoising
    & U-Net-S & $10^{-3}$ & 57,330
    & U-Net-S & $10^{-4}$ & 408,000
    & cEBM-S & $10^{-3}$ & 1,026,558
    \\

    \hline &&&&&&&&& \\
    {\bf CelebA} &&&&&&&&& \\ 
    \hline
    Inpainting Eyes
    & U-Net-L & $10^{-4}$ & 7,555
    & U-Net-L & $10^{-5}$ & 22,500
    & cEBM-L & $10^{-4}$ & 94,000
    \\
    Colorization
    & U-Net-L & $3\cdot10^{-5}$ & 9,066
    & U-Net-L & $3\cdot10^{-5}$ & 38,500
    & cEBM-L & $10^{-4}$ & 44,500
    \\
    Super-resolution
    & EDSR & $10^{-4}$ & 287,651
    & \makecell{EDSR\\+ U-Net-L} & $10^{-5}$ & 113,500
    & \makecell{EDSR\\+ cEBM-L} & $10^{-4}$ & 203,250
    \\

    \hline &&&&&&&&& \\
    {\bf ImageNet} &&&&&&&&& \\ 
    \hline
    Colorization
    & U-Net-M & $3\cdot10^{-4}$ & 187,200
    & U-Net-M & $10^{-5}$ & 37,600
    & cEBM-L & $10^{-4}$ & 102,500
    \\
    \hline &&&&&&&&& \\
    {\bf Biological} &&&&&&&&& \\ 
    \hline
    Image-to-image
    & Pretrained & &
    & U-Net-M & $10^{-4}$ & 81,100
    & cEBM-L & $10^{-4}$ & 402,800
    \\
    \hline
\end{tabular}
\end{table*}






\section{Additional Results}

\subsection{PPDs}

\begin{figure*}
    \centering
    \includegraphics[width=0.31\linewidth]{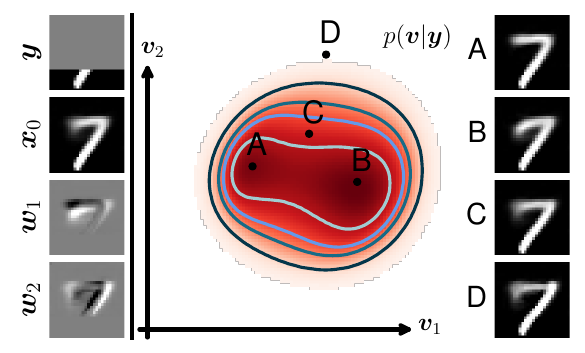}
    \includegraphics[width=0.31\linewidth]{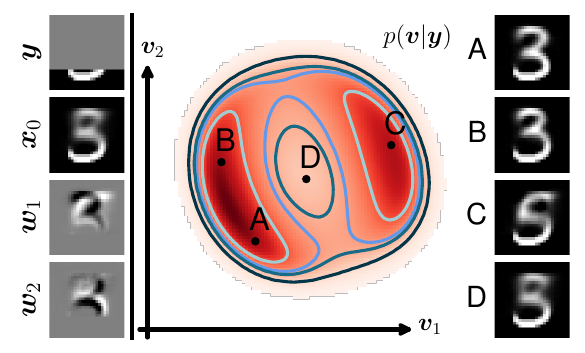}
    \includegraphics[width=0.31\linewidth]{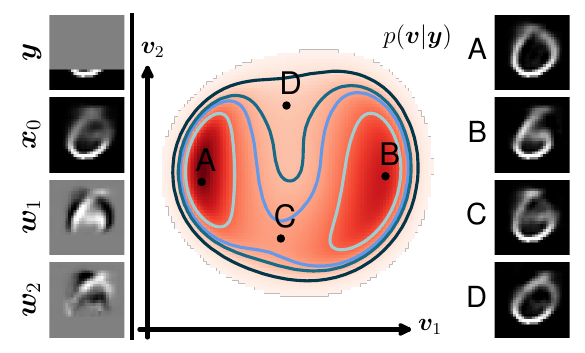}

    \includegraphics[width=0.31\linewidth]{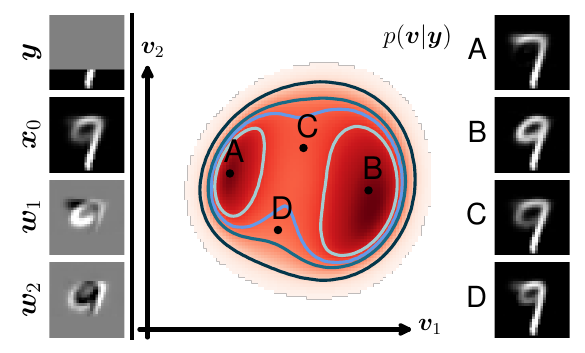}
    \includegraphics[width=0.31\linewidth]{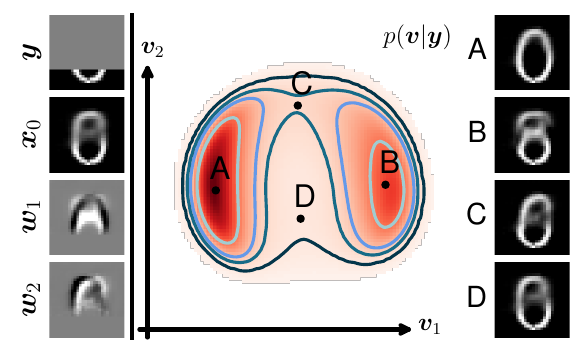}
    \includegraphics[width=0.31\linewidth]{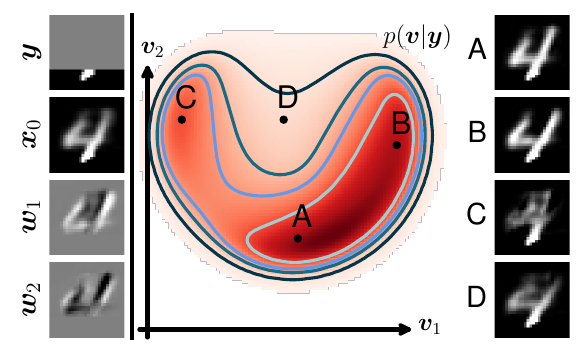}

    \includegraphics[width=0.31\linewidth]{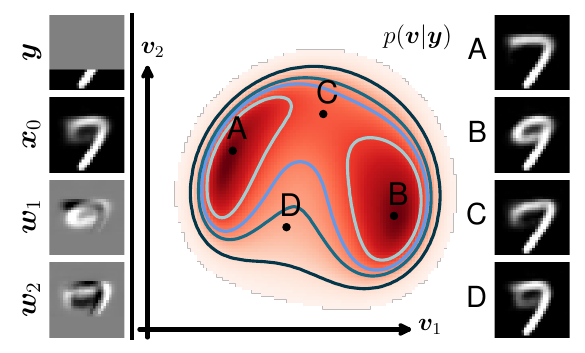}
    \includegraphics[width=0.31\linewidth]{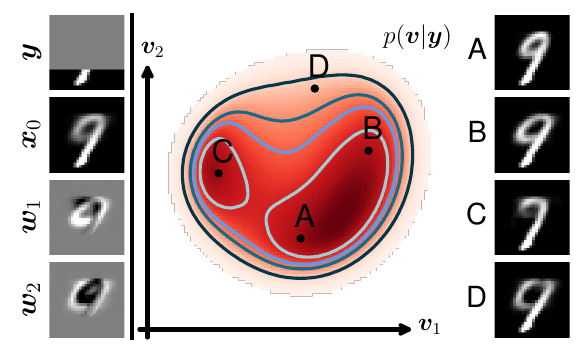}
    \includegraphics[width=0.31\linewidth]{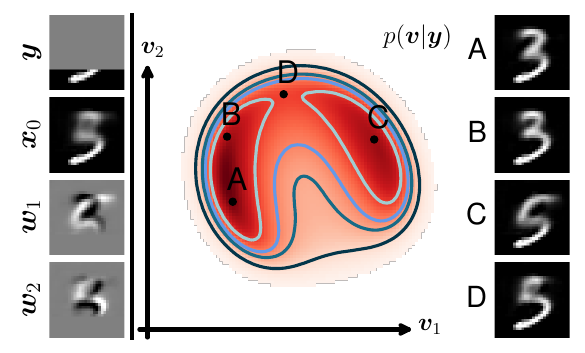}

    \caption{More examples of MNIST inpainting PPDs.}
    \label{fig:app-mnist-inpaint-ppds}
\end{figure*}

Figures~\ref{fig:app-mnist-inpaint-ppds}-\ref{fig:app-bio-ppds} provide more examples of PPDs captured by PPDE for the datasets and tasks presented in the main text.

\begin{figure*}
\centering
    \includegraphics[width=0.33\linewidth]{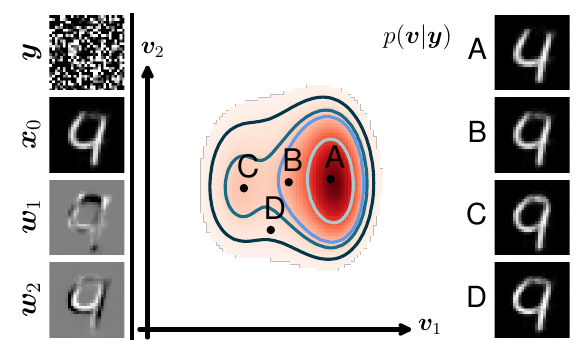}
    \includegraphics[width=0.33\linewidth]{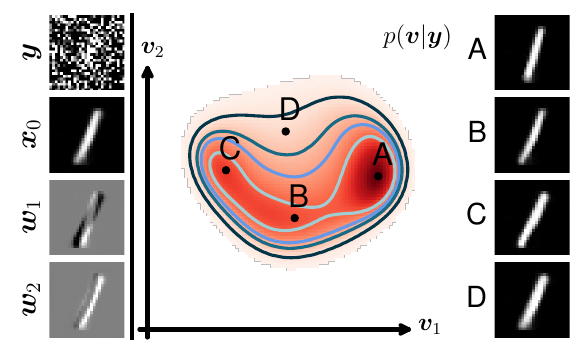}
    \includegraphics[width=0.33\linewidth]{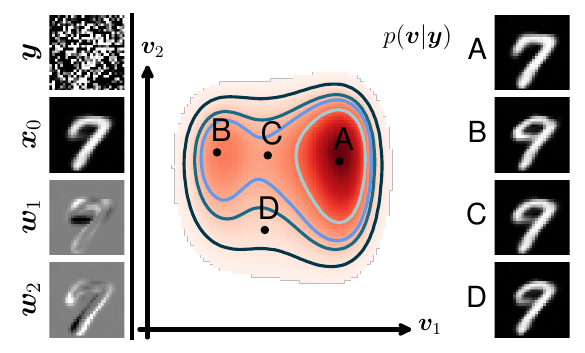}
    
    \includegraphics[width=0.33\linewidth]{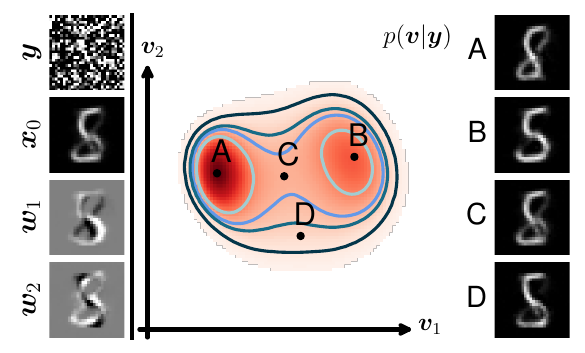}
    \includegraphics[width=0.33\linewidth]{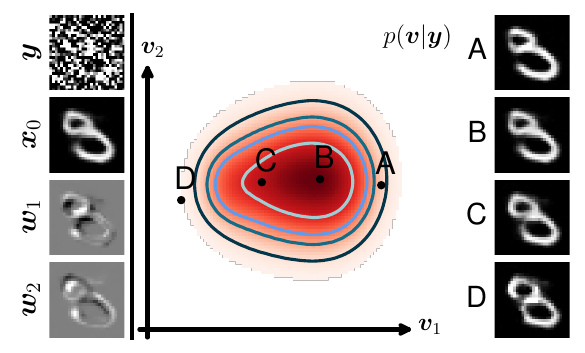}
    \includegraphics[width=0.33\linewidth]{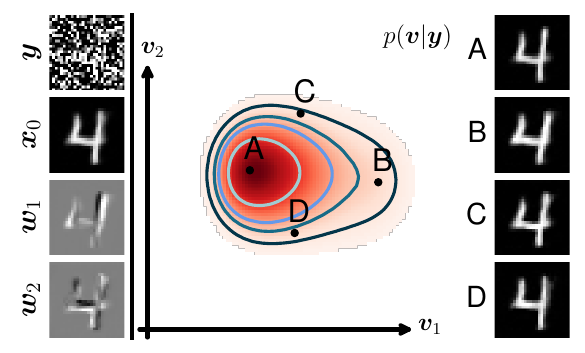}
    
    \includegraphics[width=0.33\linewidth]{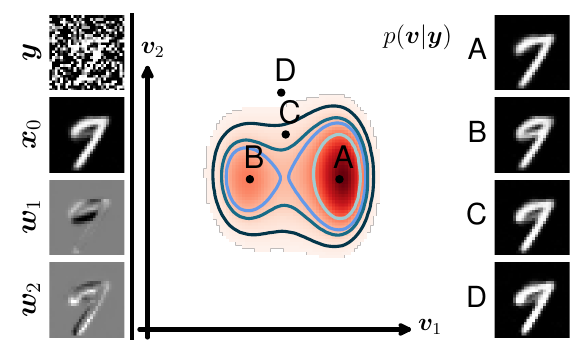}
    \includegraphics[width=0.33\linewidth]{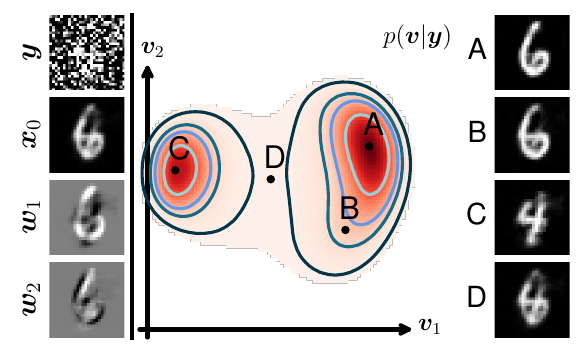}
    \includegraphics[width=0.33\linewidth]{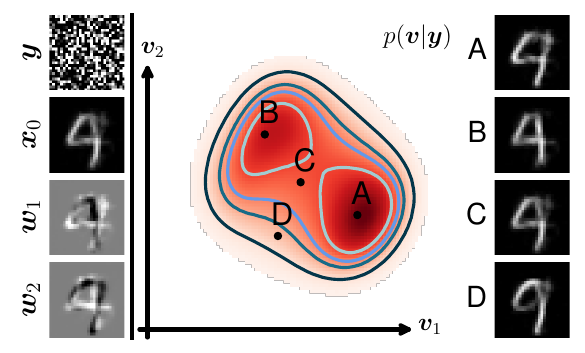}

    \caption{More examples of MNIST denoising.}
\end{figure*}

\begin{figure*}
    \centering
    \includegraphics[width=0.48\linewidth]{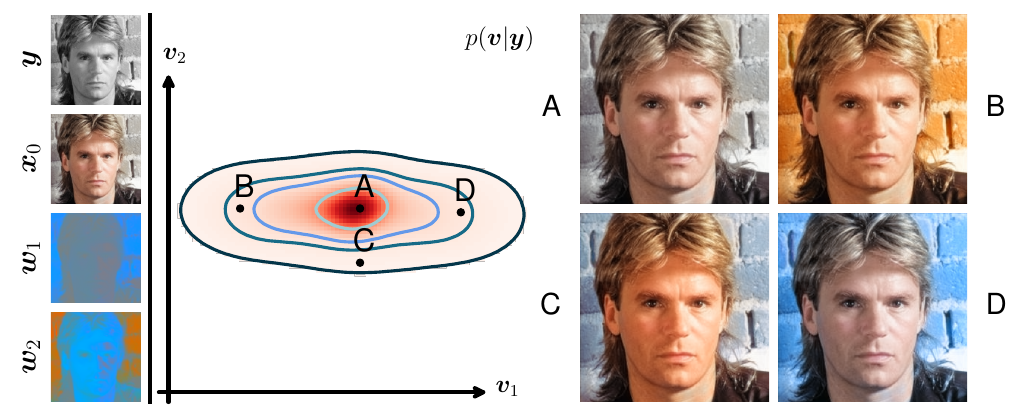}
    \includegraphics[width=0.48\linewidth]{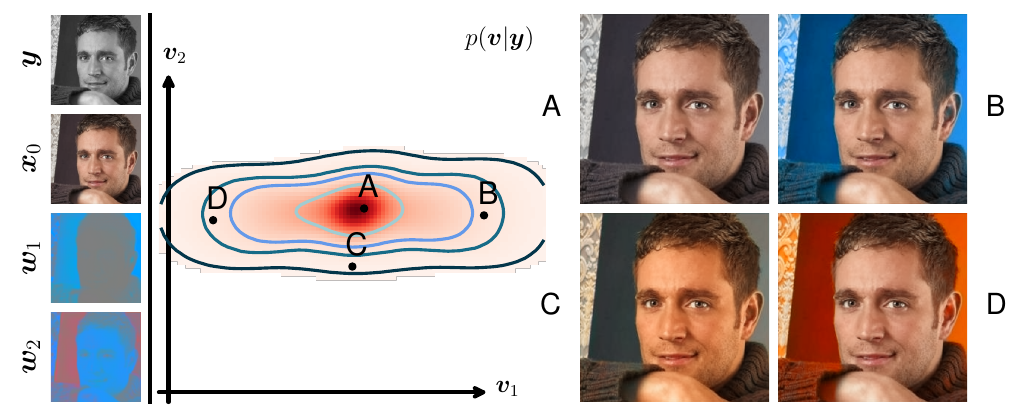}

    \includegraphics[width=0.48\linewidth]{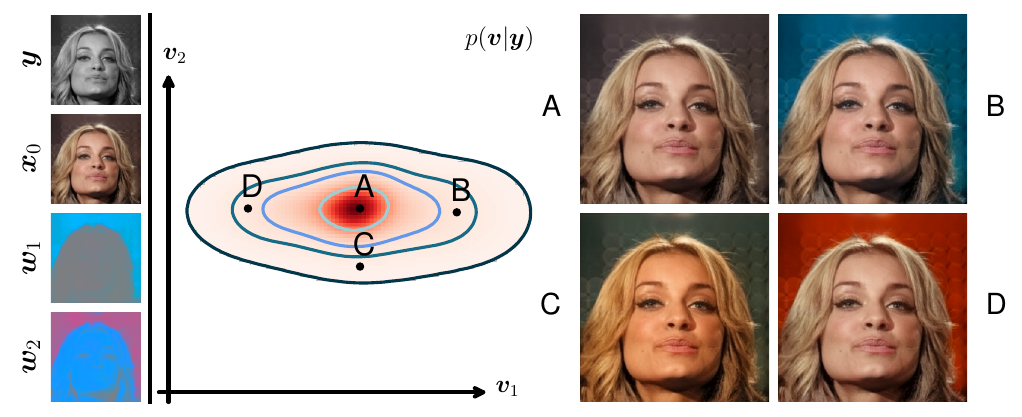}
    \includegraphics[width=0.48\linewidth]{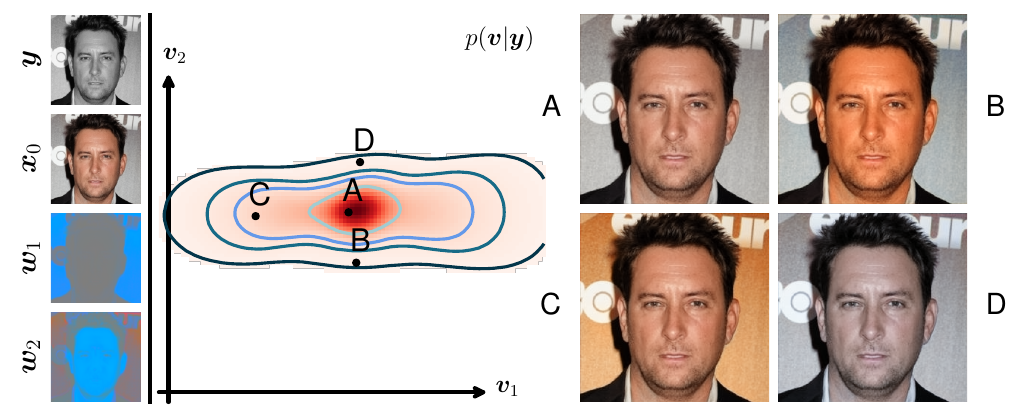}

    \includegraphics[width=0.48\linewidth]{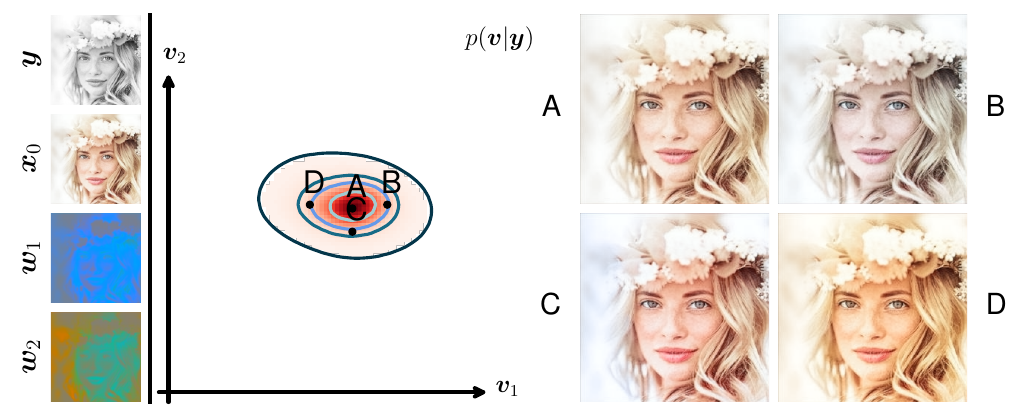}
    \includegraphics[width=0.48\linewidth]{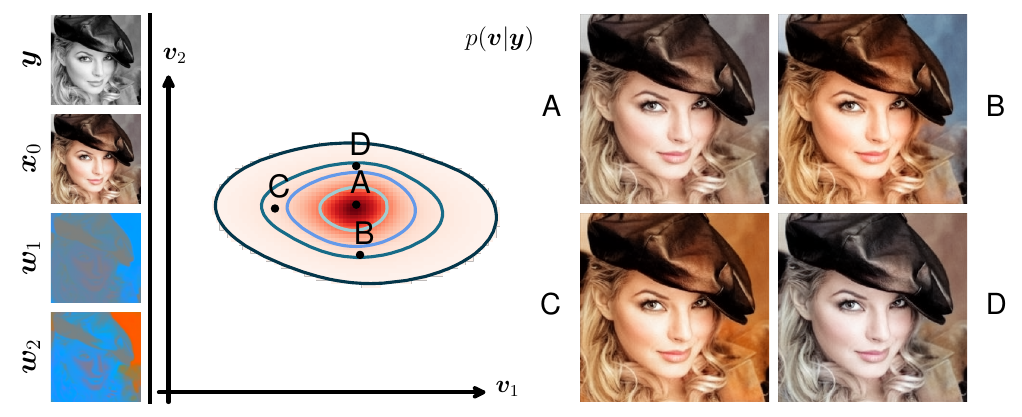}

    \caption{More examples of CelebA-HQ colorization.}
    \label{fig:celeba-colorization}
\end{figure*}

\begin{figure*}
    \centering
    \includegraphics[width=0.48\linewidth]{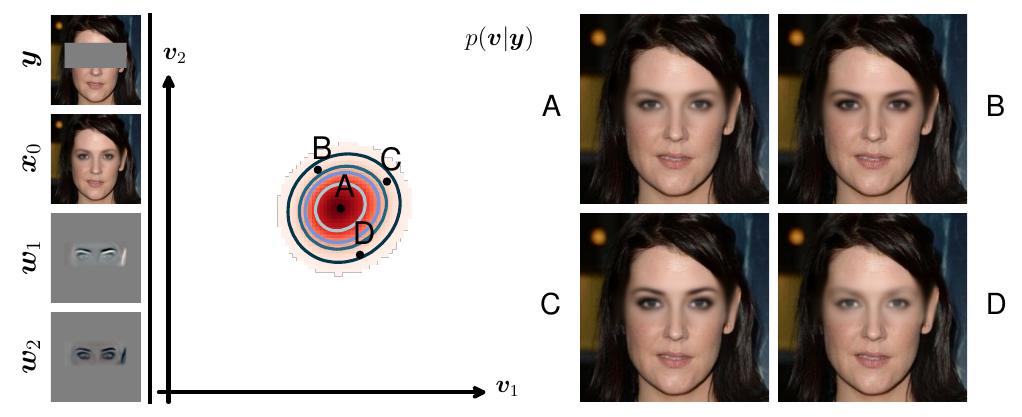}
    \includegraphics[width=0.48\linewidth]{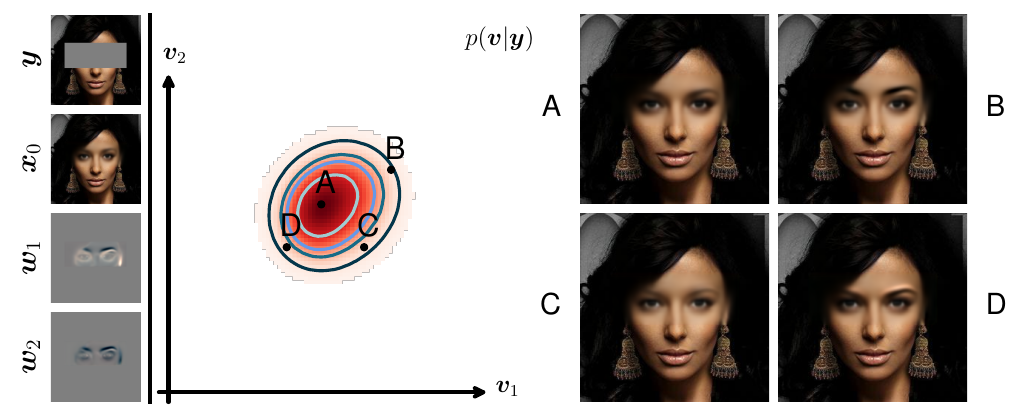}
    
    \includegraphics[width=0.48\linewidth]{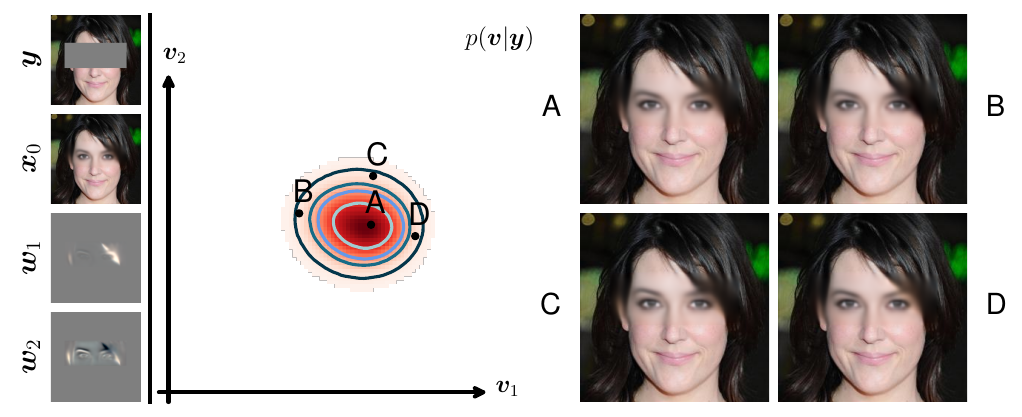}
    \includegraphics[width=0.48\linewidth]{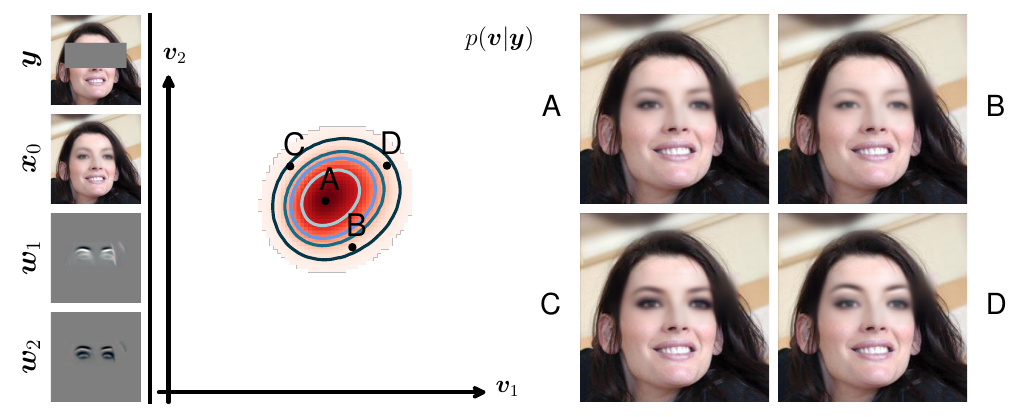}
    
    \includegraphics[width=0.48\linewidth]{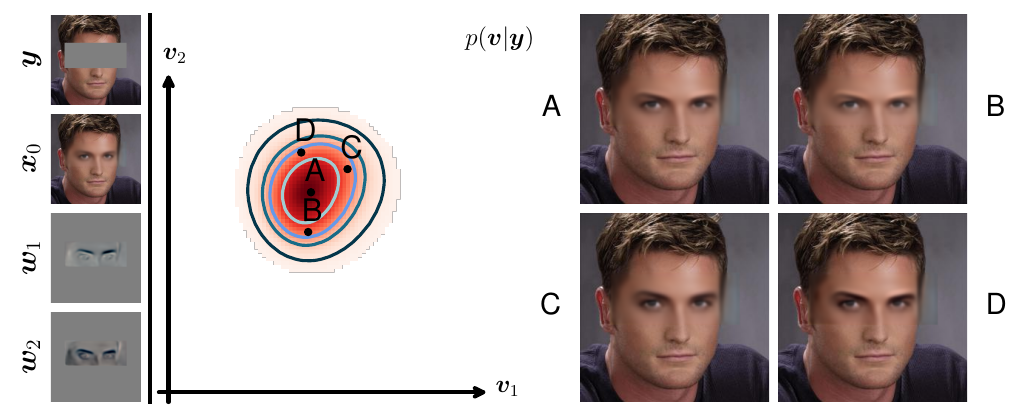}
    \includegraphics[width=0.48\linewidth]{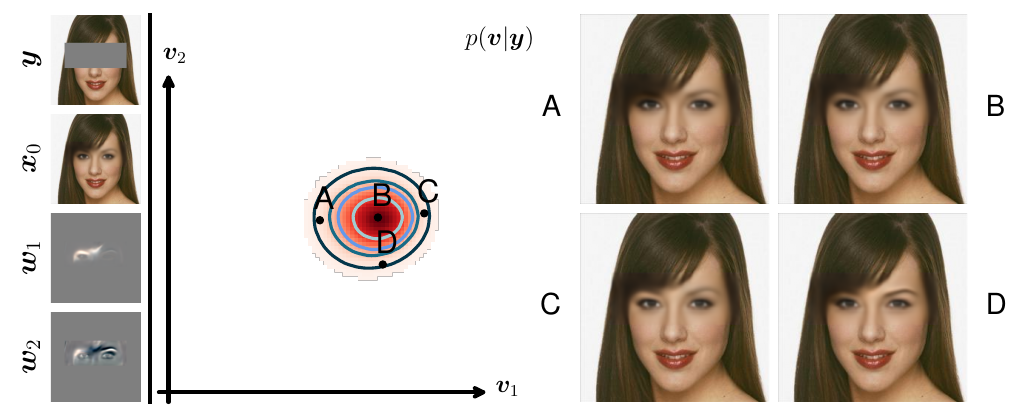}

    \caption{More examples of CelebA-HQ inpainting.}
    \label{fig:celeba-inpainting}
\end{figure*}

\begin{figure*}
    \centering

    \includegraphics[width=0.48\linewidth]{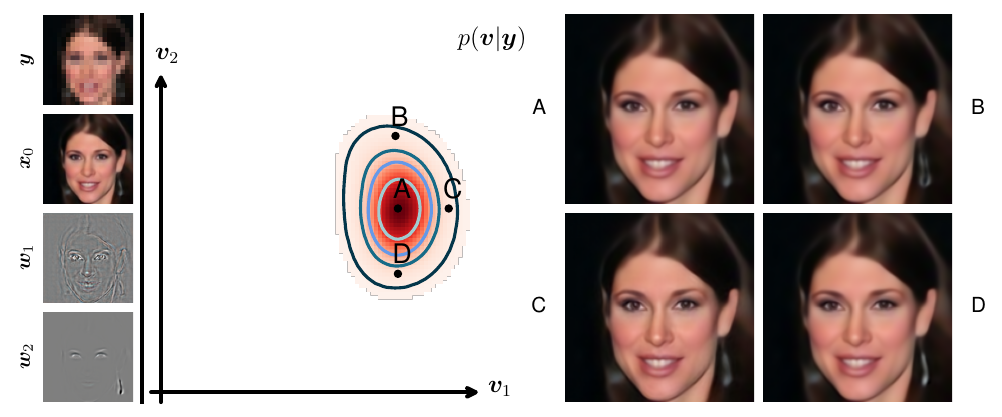}
    \includegraphics[width=0.48\linewidth]{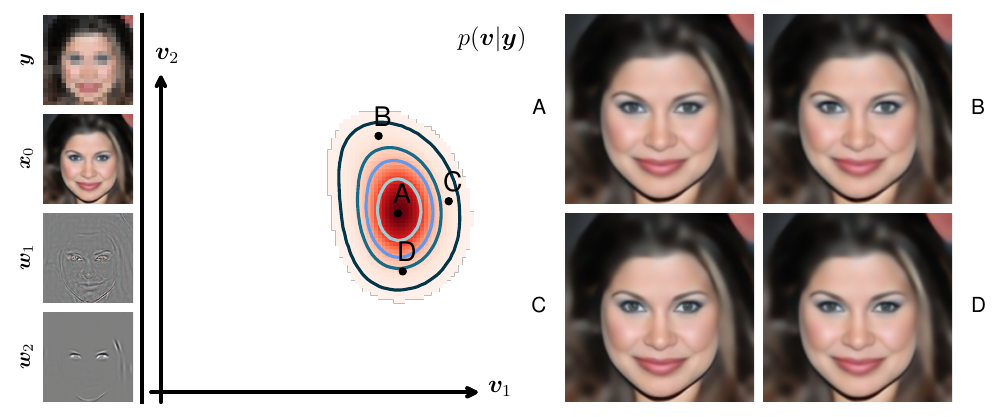}

    \includegraphics[width=0.48\linewidth]{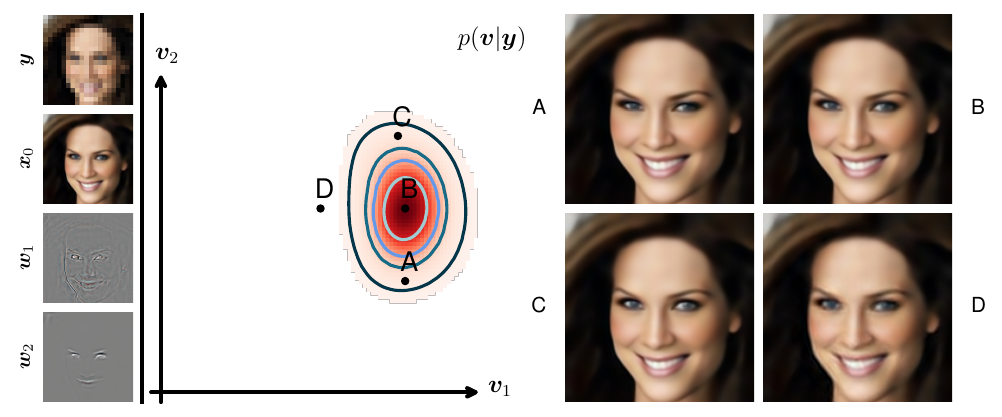}
    \includegraphics[width=0.48\linewidth]{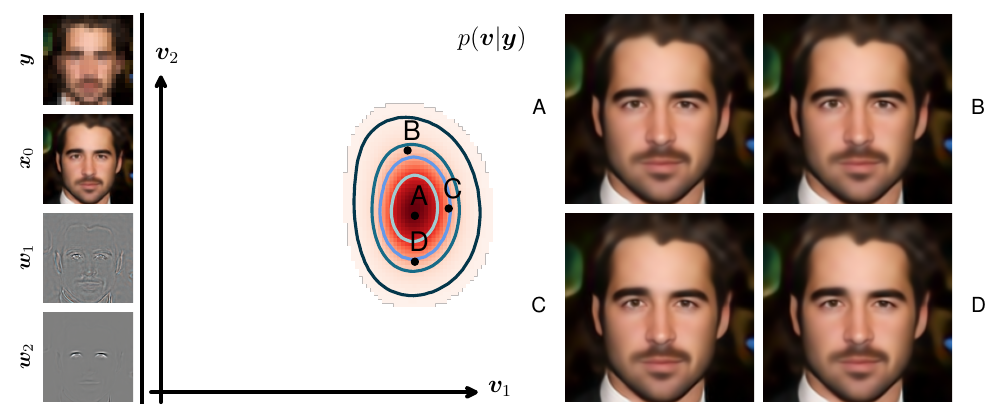}
    \caption{More examples of CelebA 8$\times$ super-resolution.}
    \label{fig:celeba-super-resolution}
\end{figure*}

\begin{figure*}
    \centering
    \includegraphics[width=0.48\linewidth]{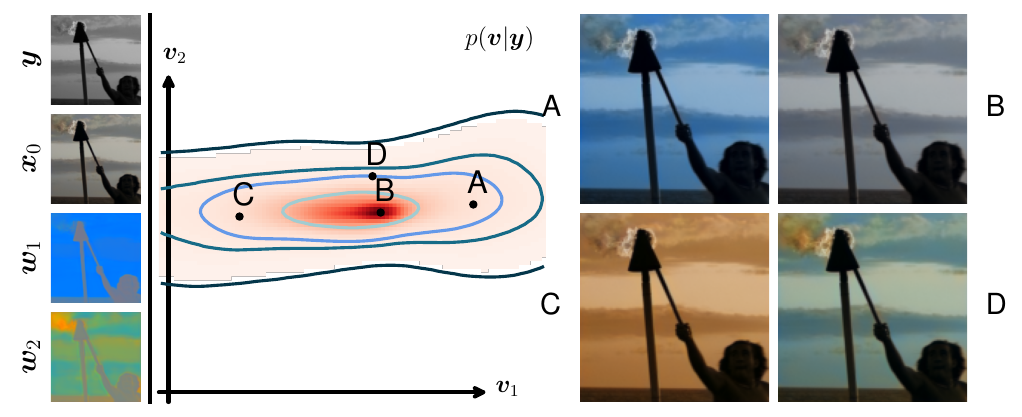}
    \includegraphics[width=0.48\linewidth]{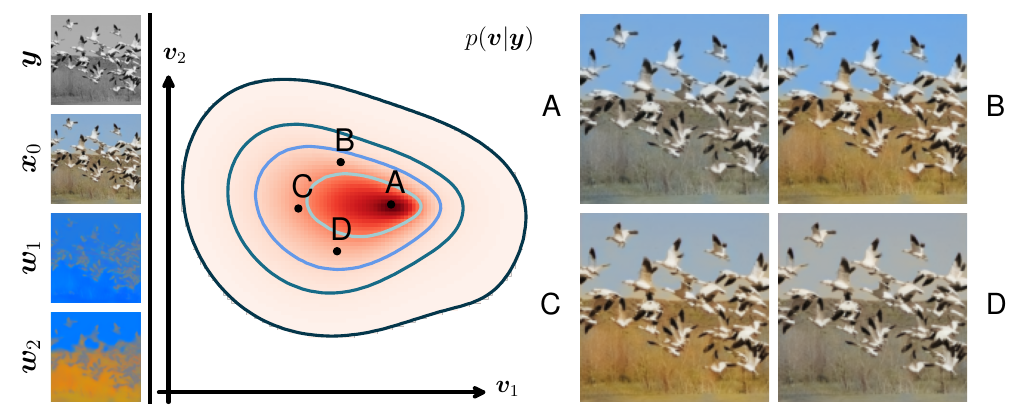}

    \includegraphics[width=0.48\linewidth]{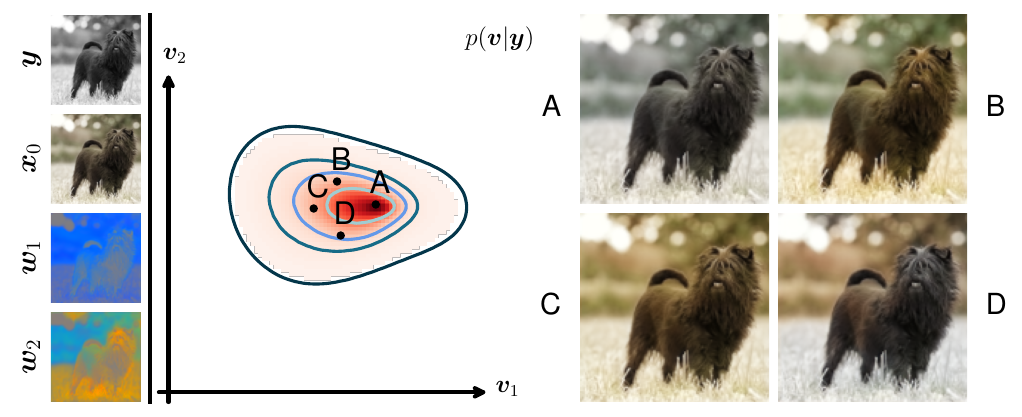}
    \includegraphics[width=0.48\linewidth]{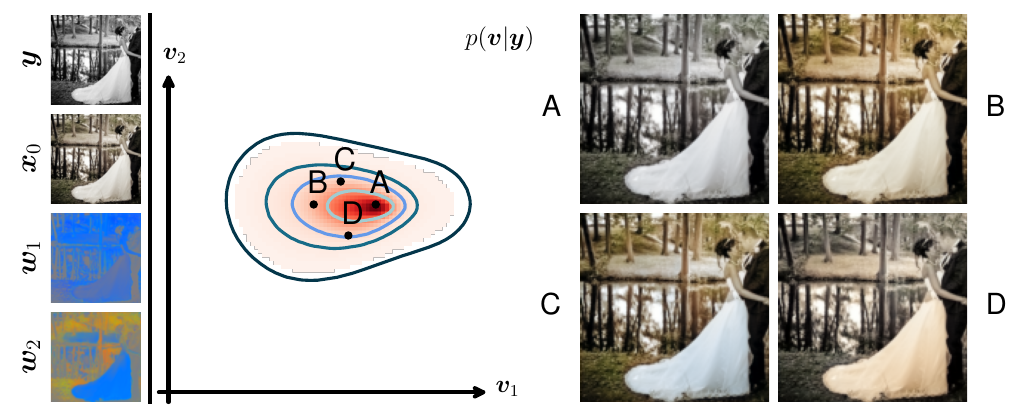}

    \includegraphics[width=0.48\linewidth]{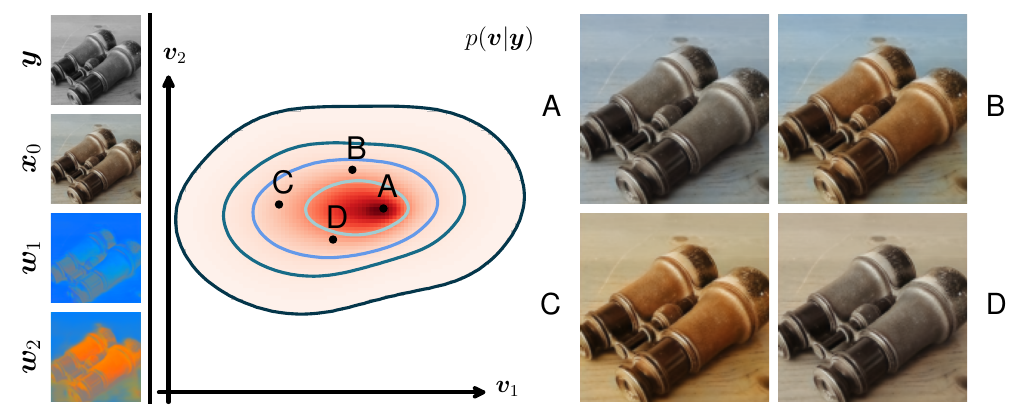}
    \includegraphics[width=0.48\linewidth]{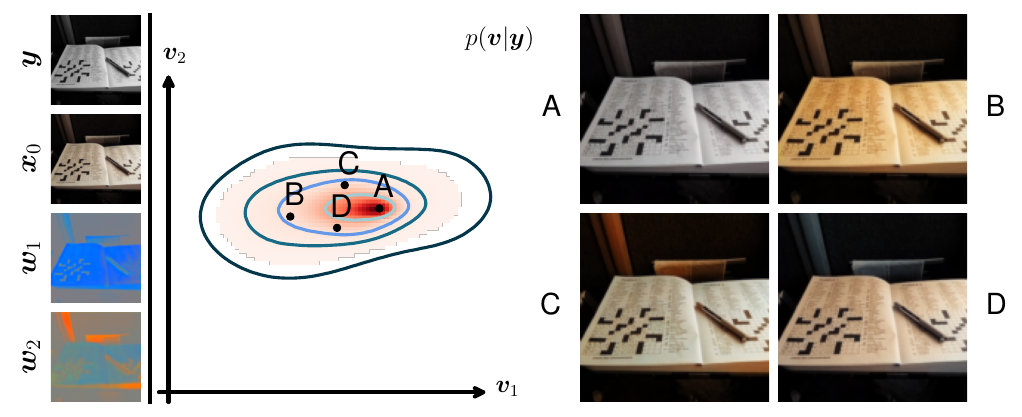}

    \caption{More examples of ImageNet colorizaion.}
    \label{fig:app-imagenet-col-ppds}
\end{figure*}

\begin{figure*}
    \centering
    \includegraphics[width=0.6\linewidth]{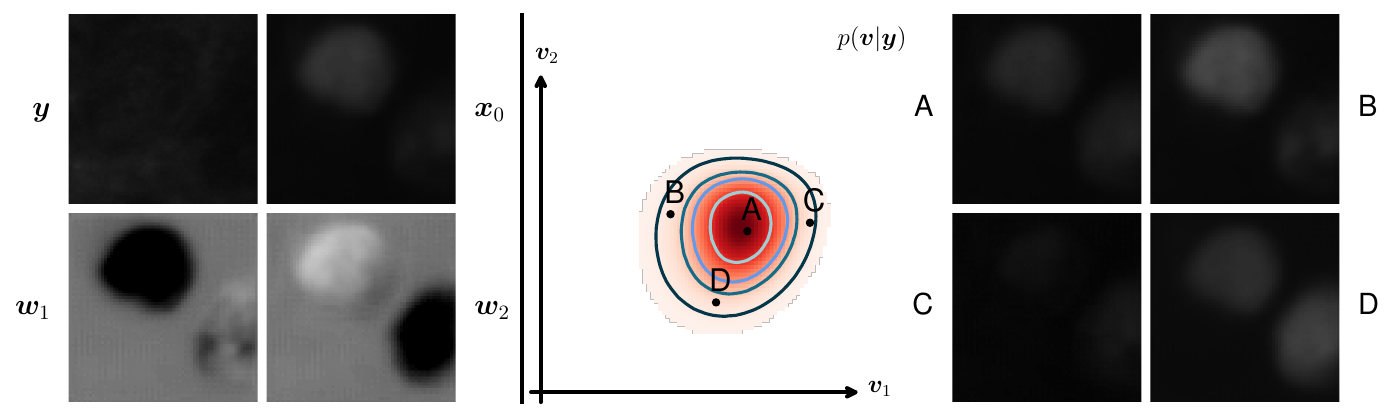}
    \includegraphics[width=0.6\linewidth]{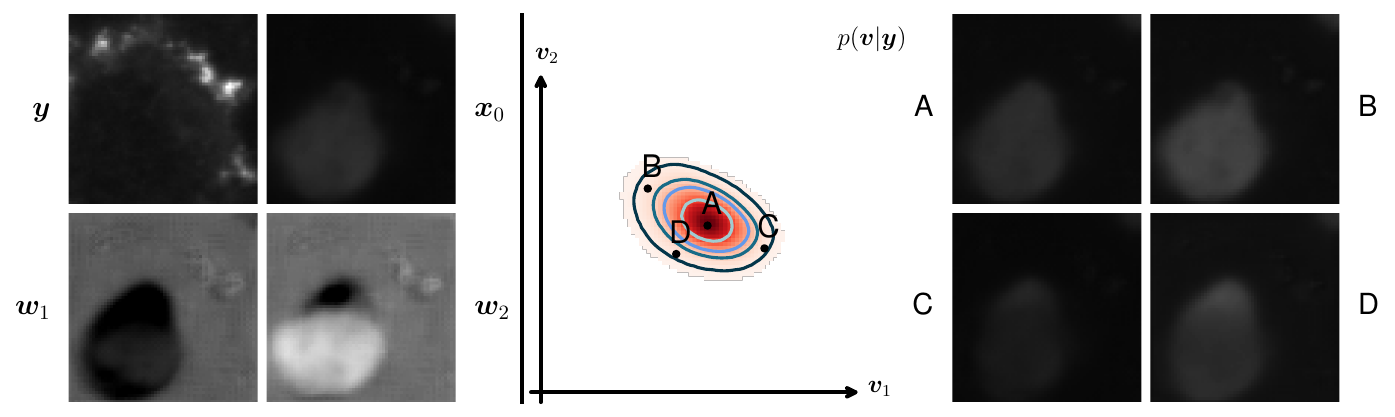}
    \includegraphics[width=0.6\linewidth]{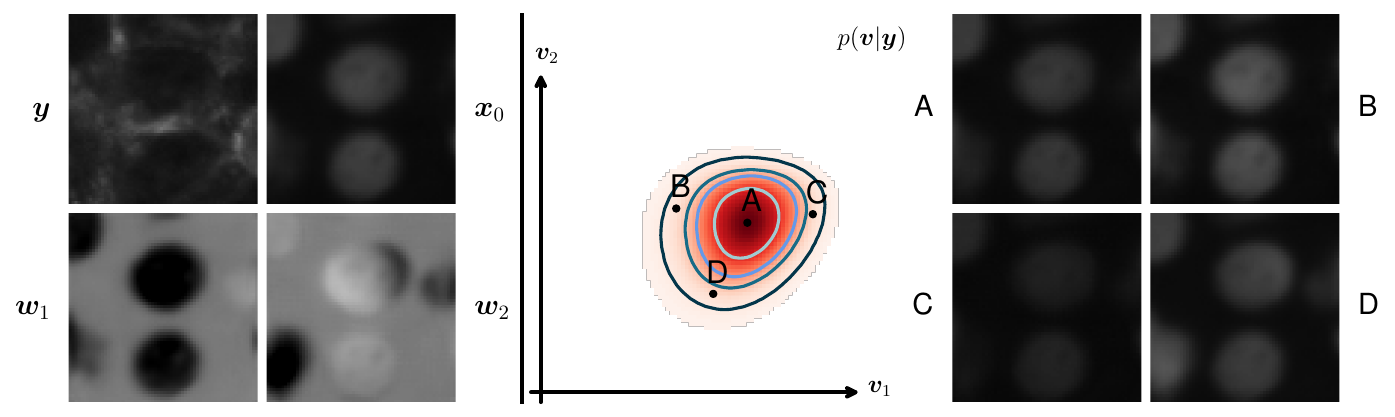}
    \includegraphics[width=0.6\linewidth]{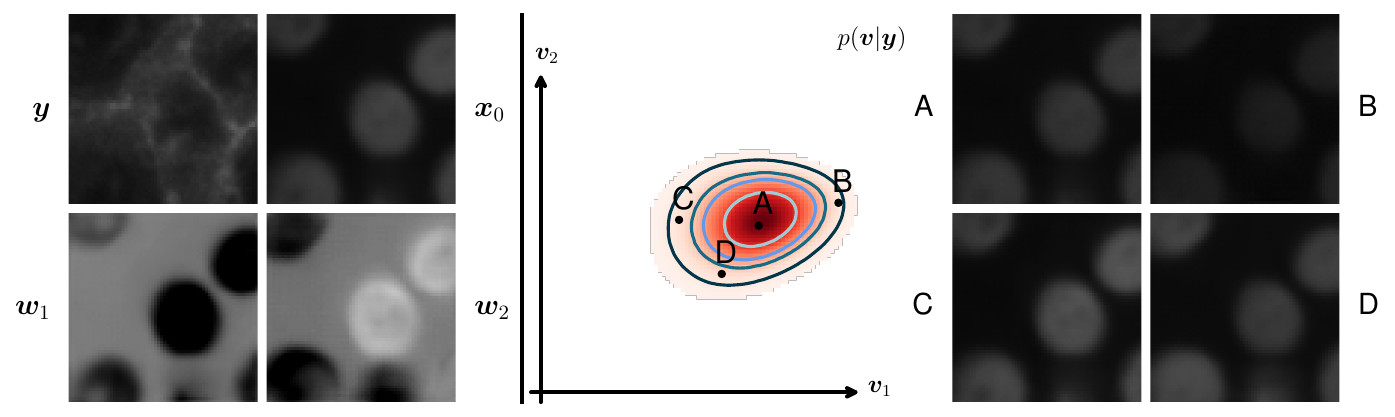}

    \caption{More examples of Biological image-to-image transfer.}
    \label{fig:app-bio-ppds}
\end{figure*}


\subsection{Comparisons to NPPC and KDE}

\begin{figure*}
    \centering
    \includegraphics[width=0.95\linewidth]{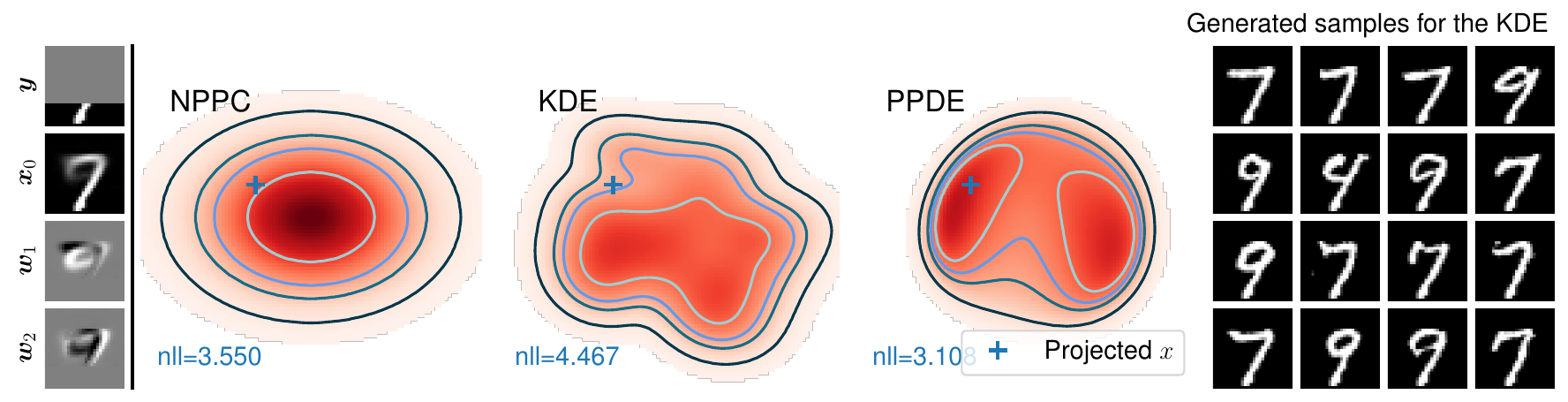}
    
    \includegraphics[width=0.95\linewidth]{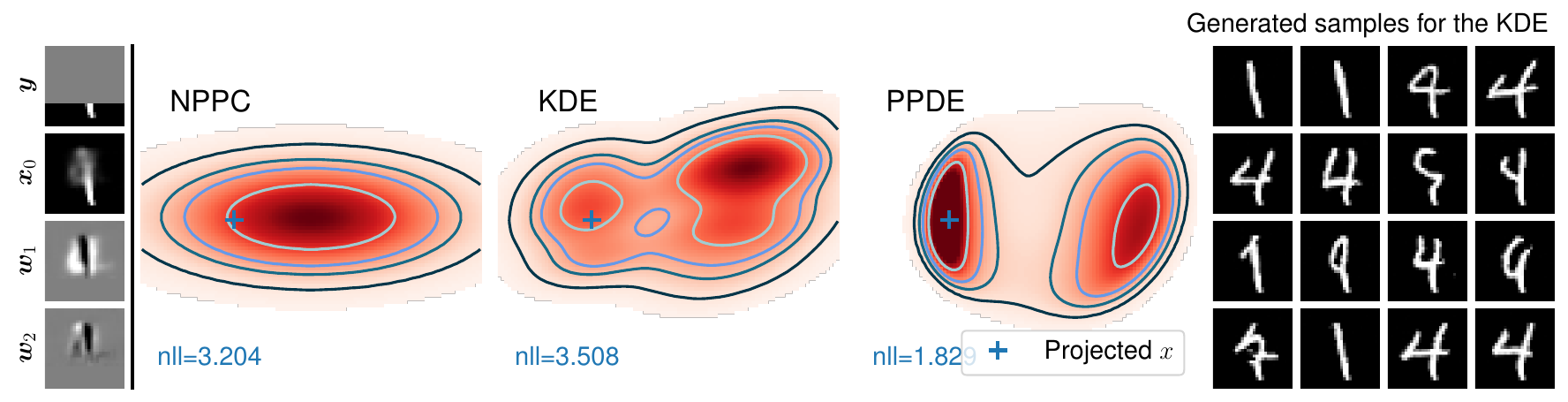}    
    \caption{More comparisons of MNIST inpainting PPDs. The posterior samples on the right were obtained using an EBM trained on MNIST.}
    \label{fig:app-mnist-inpaint-comp}
\end{figure*}

\begin{figure*}
    \centering
    \includegraphics[width=0.95\linewidth]{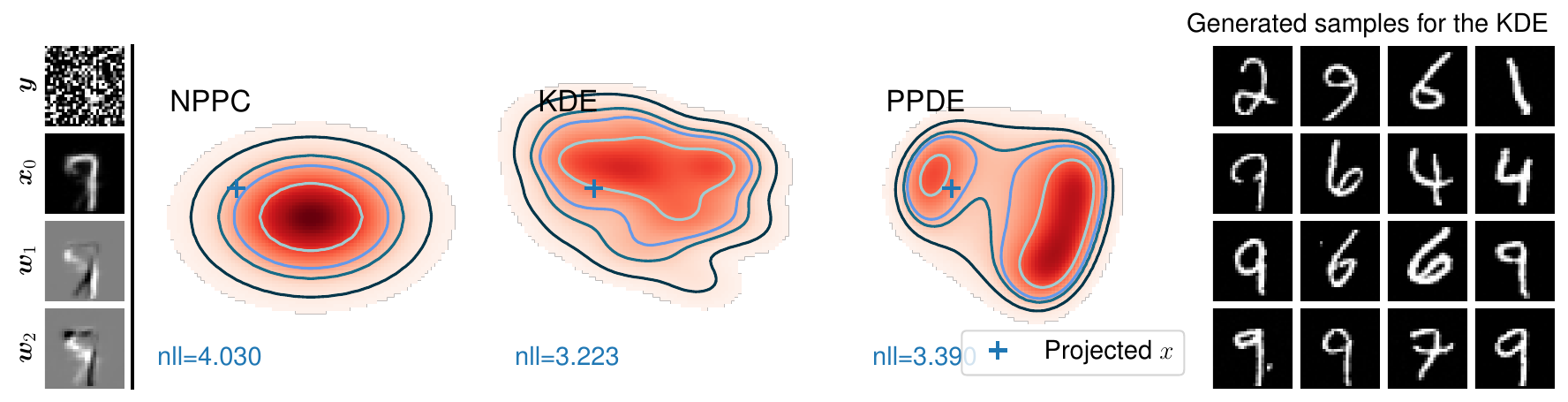}
    
    \includegraphics[width=0.95\linewidth]{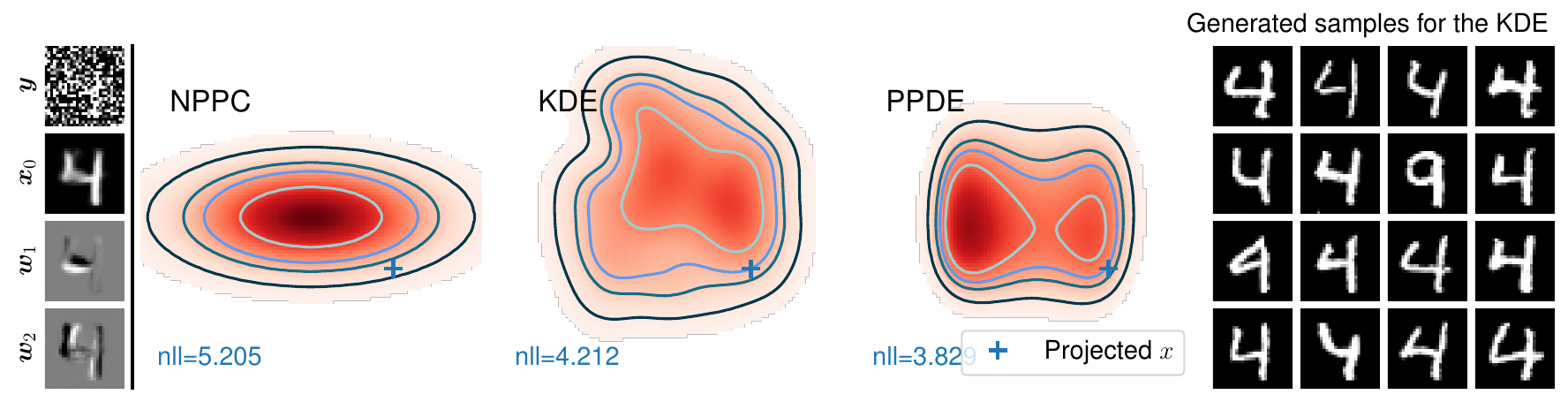}    
    \caption{More comparisons of MNIST denoising. The posterior samples on the right were obtained using an EBM trained on MNIST.}
    \label{fig:app-mnist-denoise-comp}
\end{figure*}

\begin{figure*}
    \centering
    \includegraphics[width=0.95\linewidth]{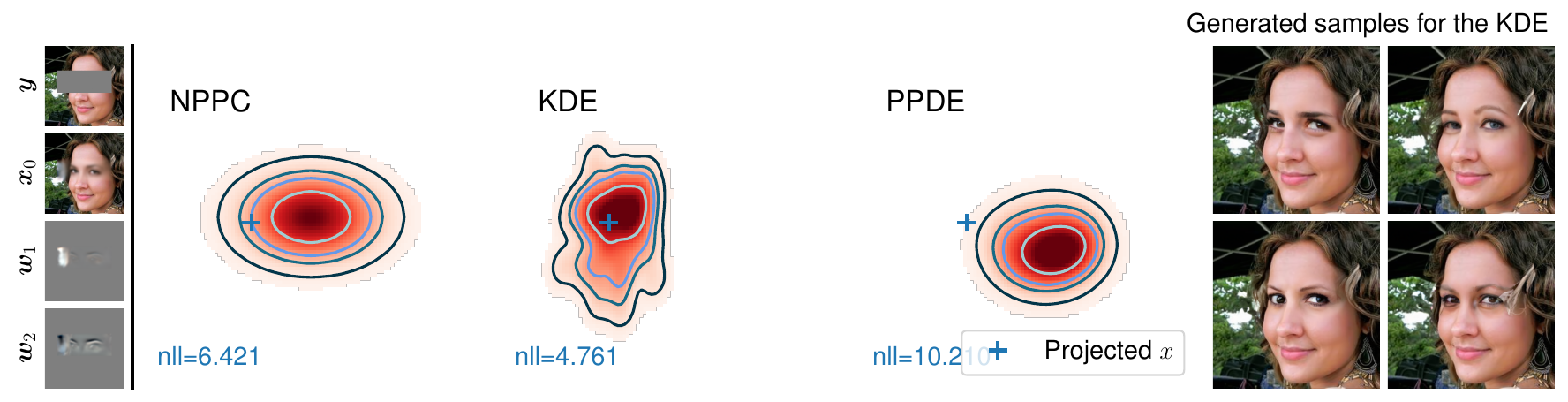}

    \includegraphics[width=0.95\linewidth]{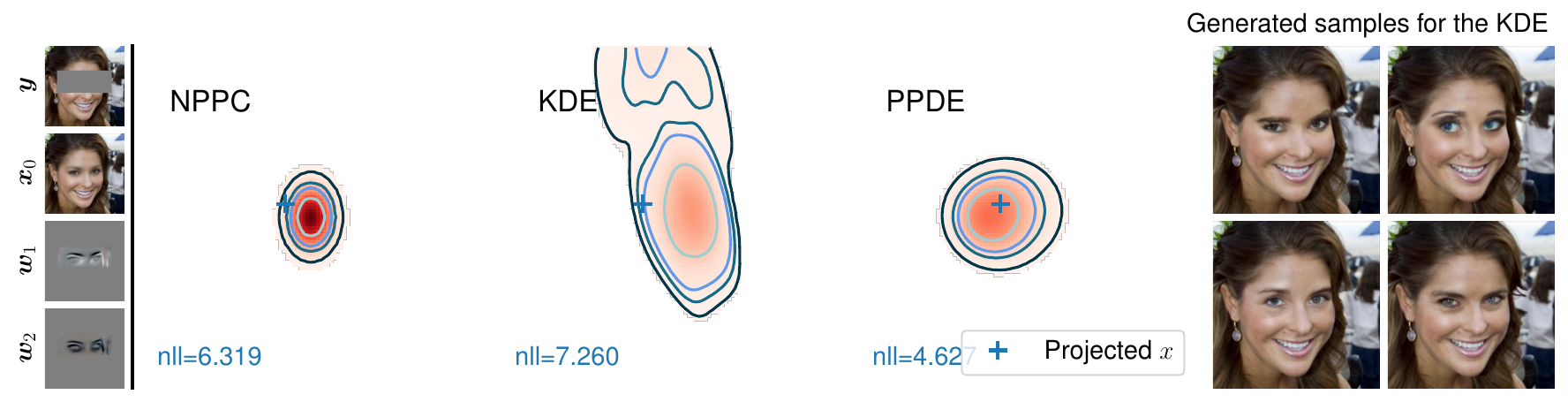}
    \caption{More comparisons of CelebA-HQ inpainting. The posterior samples on the right were obtained using DPS \citep{li2022mat}.}
    \label{fig:app-celebahq-inpaint-comp}
\end{figure*}

\begin{figure*}
    \centering
    \includegraphics[width=0.95\linewidth]{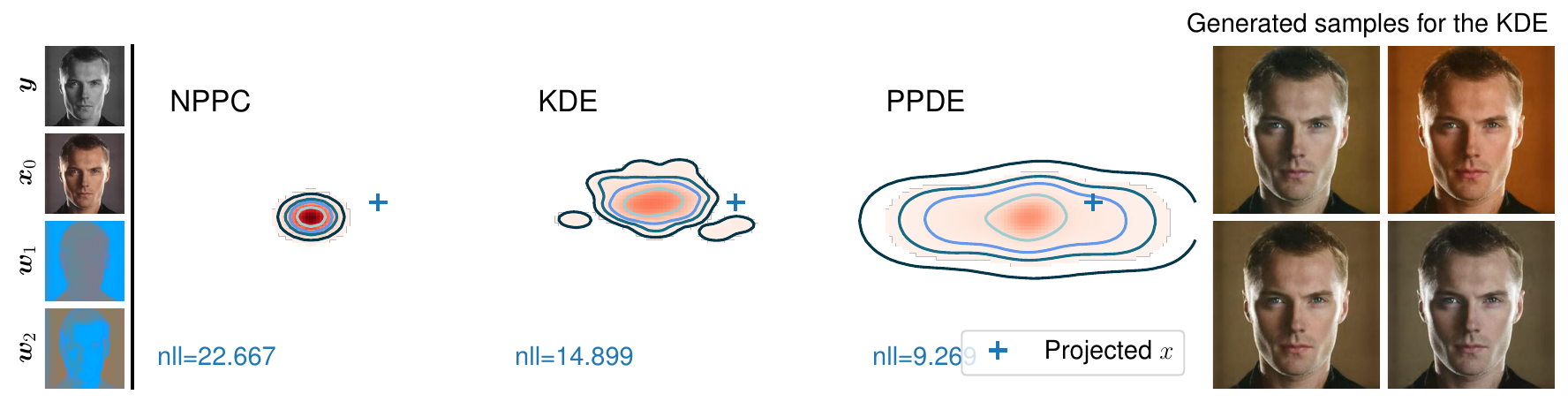}

    \includegraphics[width=0.95\linewidth]{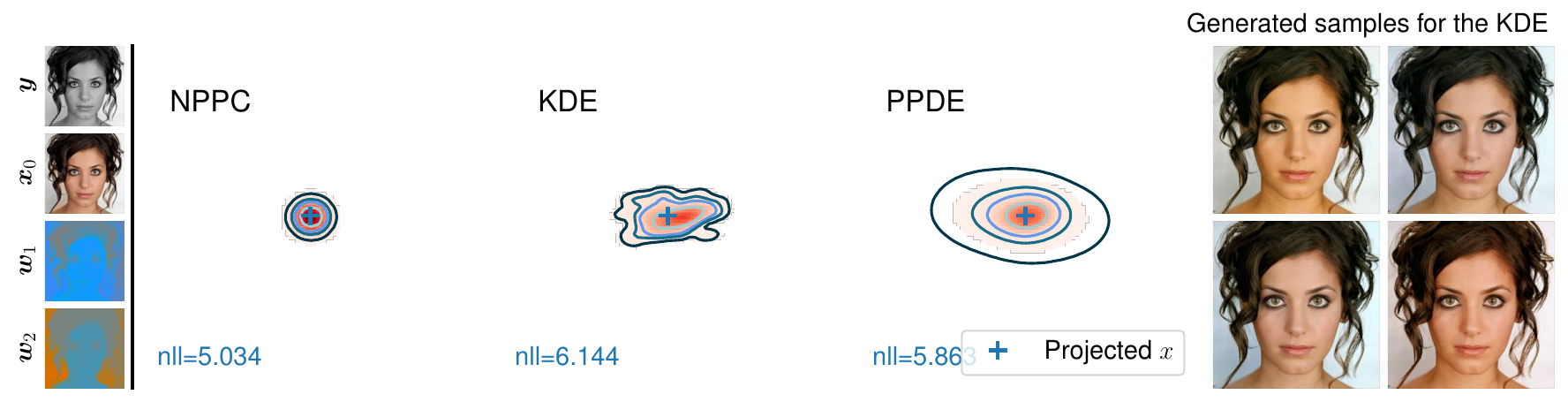}
    \caption{More comparisons of CelebA-HQ colorization. The posterior samples on the right were obtained using DDNM \citep{wang2023ddnm}.}
    \label{fig:app-celebahq-col-comp}
\end{figure*}

\begin{figure*}
    \centering

    \includegraphics[width=0.65\linewidth]{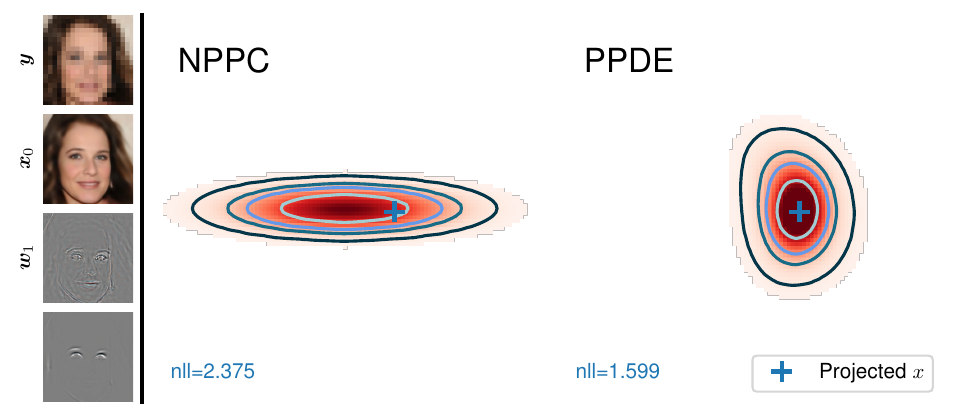}
    \caption{More comparisons of CelebA 8$\times$ super-resolution.}
    \label{fig:app-celeba-superres-comp}
\end{figure*}

\begin{figure*}
    \centering

    \includegraphics[width=0.65\linewidth]{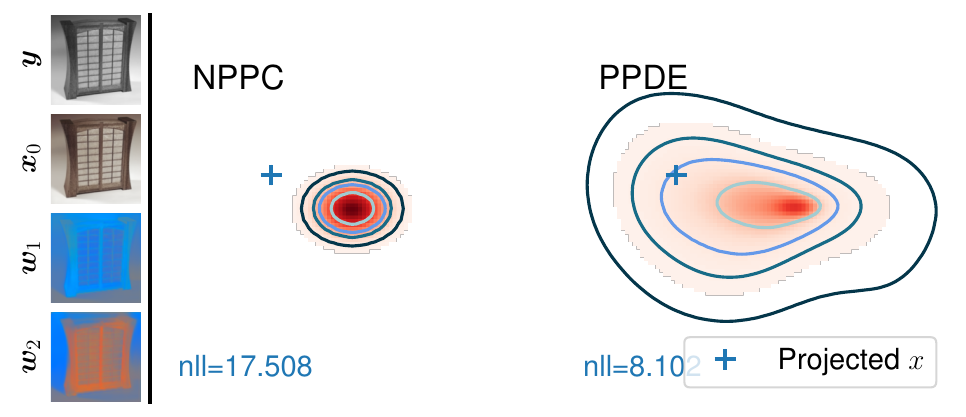}
    \caption{More comparisons of ImageNet colorization.}
    \label{fig:app-imagenet-col-comp}
\end{figure*}

Figures~\ref{fig:app-mnist-inpaint-comp}-\ref{fig:app-imagenet-col-comp} provide further negative log-likelihood comparisons of PPDE with the respective baselines.


\clearpage
{
    \small
    \bibliographystyle{ieeenat_fullname}
    \bibliography{bib}

\begin{thebibliography}{76}
\providecommand{\natexlab}[1]{#1}
\providecommand{\url}[1]{\texttt{#1}}
\expandafter\ifx\csname urlstyle\endcsname\relax
  \providecommand{\doi}[1]{doi: #1}\else
  \providecommand{\doi}{doi: \begingroup \urlstyle{rm}\Url}\fi

\bibitem[Amini et~al.(2020)Amini, Schwarting, Soleimany, and
  Rus]{amini2020deep}
Alexander Amini, Wilko Schwarting, Ava Soleimany, and Daniela Rus.
\newblock Deep evidential regression.
\newblock \emph{Advances in Neural Information Processing Systems},
  33:\penalty0 14927--14937, 2020.

\bibitem[Angelopoulos et~al.(2022)Angelopoulos, Kohli, Bates, Jordan, Malik,
  Alshaabi, Upadhyayula, and Romano]{angelopoulos2022image}
Anastasios~N Angelopoulos, Amit~P Kohli, Stephen Bates, Michael~I Jordan,
  Jitendra Malik, Thayer Alshaabi, Srigokul Upadhyayula, and Yaniv Romano.
\newblock Image-to-image regression with distribution-free uncertainty
  quantification and applications in imaging.
\newblock In \emph{International Conference on Machine Learning}, 2022.

\bibitem[Bates et~al.(2021)Bates, Angelopoulos, Lei, Malik, and
  Jordan]{bates2021distribution}
Stephen Bates, Anastasios Angelopoulos, Lihua Lei, Jitendra Malik, and Michael
  Jordan.
\newblock Distribution-free, risk-controlling prediction sets.
\newblock \emph{Journal of the ACM (JACM)}, 68\penalty0 (6):\penalty0 1--34,
  2021.

\bibitem[Belhasin et~al.(2023)Belhasin, Romano, Freedman, Rivlin, and
  Elad]{belhasin2023principal}
Omer Belhasin, Yaniv Romano, Daniel Freedman, Ehud Rivlin, and Michael Elad.
\newblock Principal uncertainty quantification with spatial correlation for
  image restoration problems.
\newblock \emph{arXiv preprint arXiv:2305.10124}, 2023.

\bibitem[Bendel et~al.(2022)Bendel, Ahmad, and Schniter]{bendel2022regularized}
Matthew Bendel, Rizwan Ahmad, and Philip Schniter.
\newblock A regularized conditional gan for posterior sampling in inverse
  problems.
\newblock \emph{arXiv preprint arXiv:2210.13389}, 2022.

\bibitem[Bengio and Delalleau(2009)]{bengio2009justifying}
Yoshua Bengio and Olivier Delalleau.
\newblock Justifying and generalizing contrastive divergence.
\newblock \emph{Neural computation}, 21\penalty0 (6):\penalty0 1601--1621,
  2009.

\bibitem[Blundell et~al.(2015)Blundell, Cornebise, Kavukcuoglu, and
  Wierstra]{blundell2015weight}
Charles Blundell, Julien Cornebise, Koray Kavukcuoglu, and Daan Wierstra.
\newblock Weight uncertainty in neural network.
\newblock In \emph{International conference on machine learning}, pages
  1613--1622. PMLR, 2015.

\bibitem[Carreira-Perpinan and Hinton(2005)]{carreira2005contrastive}
Miguel~A Carreira-Perpinan and Geoffrey Hinton.
\newblock On contrastive divergence learning.
\newblock In \emph{International workshop on artificial intelligence and
  statistics}, pages 33--40. PMLR, 2005.

\bibitem[Christiansen et~al.(2018)Christiansen, Yang, Ando, Javaherian,
  Skibinski, Lipnick, Mount, O’neil, Shah, Lee,
  et~al.]{christiansen2018silico}
Eric~M Christiansen, Samuel~J Yang, D~Michael Ando, Ashkan Javaherian, Gaia
  Skibinski, Scott Lipnick, Elliot Mount, Alison O’neil, Kevan Shah, Alicia~K
  Lee, et~al.
\newblock In silico labeling: predicting fluorescent labels in unlabeled
  images.
\newblock \emph{Cell}, 173\penalty0 (3):\penalty0 792--803, 2018.

\bibitem[Chung et~al.(2022)Chung, Kim, Mccann, Klasky, and
  Ye]{chung2022diffusion}
Hyungjin Chung, Jeongsol Kim, Michael~Thompson Mccann, Marc~Louis Klasky, and
  Jong~Chul Ye.
\newblock Diffusion posterior sampling for general noisy inverse problems.
\newblock In \emph{The Eleventh International Conference on Learning
  Representations}, 2022.

\bibitem[Chung et~al.(2023)Chung, Lee, and Ye]{chung2023fast}
Hyungjin Chung, Suhyeon Lee, and Jong~Chul Ye.
\newblock Fast diffusion sampler for inverse problems by geometric
  decomposition.
\newblock \emph{arXiv preprint arXiv:2303.05754}, 2023.

\bibitem[Cohen et~al.(2023)Cohen, Manor, Bahat, and
  Michaeli]{cohen2023posterior}
Noa Cohen, Hila Manor, Yuval Bahat, and Tomer Michaeli.
\newblock From posterior sampling to meaningful diversity in image restoration.
\newblock \emph{arXiv preprint arXiv:2310.16047}, 2023.

\bibitem[Deng(2012)]{deng2012mnist}
Li Deng.
\newblock The mnist database of handwritten digit images for machine learning
  research.
\newblock \emph{IEEE Signal Processing Magazine}, 29\penalty0 (6):\penalty0
  141--142, 2012.

\bibitem[Dorta et~al.(2018)Dorta, Vicente, Agapito, Campbell, and
  Simpson]{dorta2018structured}
Garoe Dorta, Sara Vicente, Lourdes Agapito, Neill~DF Campbell, and Ivor
  Simpson.
\newblock Structured uncertainty prediction networks.
\newblock In \emph{Proceedings of the IEEE conference on computer vision and
  pattern recognition}, pages 5477--5485, 2018.

\bibitem[Du et~al.(2021)Du, Li, Tenenbaum, and Mordatch]{du2021improved}
Yilun Du, Shuang Li, Joshua~B Tenenbaum, and Igor Mordatch.
\newblock Improved contrastive divergence training of energy based models.
\newblock In \emph{Energy Based Models Workshop-ICLR 2021}, 2021.

\bibitem[Falk et~al.(2019)Falk, Mai, Bensch, {\c{C}}i{\c{c}}ek, Abdulkadir,
  Marrakchi, B{\"o}hm, Deubner, J{\"a}ckel, Seiwald, et~al.]{falk2019u}
Thorsten Falk, Dominic Mai, Robert Bensch, {\"O}zg{\"u}n {\c{C}}i{\c{c}}ek,
  Ahmed Abdulkadir, Yassine Marrakchi, Anton B{\"o}hm, Jan Deubner, Zoe
  J{\"a}ckel, Katharina Seiwald, et~al.
\newblock U-net: deep learning for cell counting, detection, and morphometry.
\newblock \emph{Nature methods}, 16\penalty0 (1):\penalty0 67--70, 2019.

\bibitem[Gal and Ghahramani(2016)]{gal2016dropout}
Yarin Gal and Zoubin Ghahramani.
\newblock Dropout as a bayesian approximation: Representing model uncertainty
  in deep learning.
\newblock In \emph{international conference on machine learning}, pages
  1050--1059. PMLR, 2016.

\bibitem[Gao et~al.(2020)Gao, Song, Poole, Wu, and Kingma]{gao2020learning}
Ruiqi Gao, Yang Song, Ben Poole, Ying~Nian Wu, and Diederik~P Kingma.
\newblock Learning energy-based models by diffusion recovery likelihood.
\newblock In \emph{International Conference on Learning Representations}, 2020.

\bibitem[Gutmann and Hyv{\"a}rinen(2010)]{gutmann2010noise}
Michael Gutmann and Aapo Hyv{\"a}rinen.
\newblock Noise-contrastive estimation: A new estimation principle for
  unnormalized statistical models.
\newblock In \emph{Proceedings of the thirteenth international conference on
  artificial intelligence and statistics}, pages 297--304. JMLR Workshop and
  Conference Proceedings, 2010.

\bibitem[HASTINGS(1970)]{hastings1970monte}
WK HASTINGS.
\newblock Monte carlo sampling methods using markov chains and their
  applications.
\newblock \emph{Biometrika}, 57\penalty0 (1):\penalty0 97--97, 1970.

\bibitem[He et~al.(2016)He, Zhang, Ren, and Sun]{he2016deep}
Kaiming He, Xiangyu Zhang, Shaoqing Ren, and Jian Sun.
\newblock Deep residual learning for image recognition.
\newblock In \emph{Proceedings of the IEEE conference on computer vision and
  pattern recognition}, pages 770--778, 2016.

\bibitem[Hinton(2002)]{hinton2002training}
Geoffrey~E Hinton.
\newblock Training products of experts by minimizing contrastive divergence.
\newblock \emph{Neural computation}, 14\penalty0 (8):\penalty0 1771--1800,
  2002.

\bibitem[Ho et~al.(2020)Ho, Jain, and Abbeel]{ho2020denoising}
Jonathan Ho, Ajay Jain, and Pieter Abbeel.
\newblock Denoising diffusion probabilistic models.
\newblock \emph{Advances in neural information processing systems},
  33:\penalty0 6840--6851, 2020.

\bibitem[Horwitz and Hoshen(2022)]{horwitz2022conffusion}
Eliahu Horwitz and Yedid Hoshen.
\newblock Conffusion: Confidence intervals for diffusion models.
\newblock \emph{arXiv preprint arXiv:2211.09795}, 2022.

\bibitem[Huang and Belongie(2017)]{huang2017arbitrary}
Xun Huang and Serge Belongie.
\newblock Arbitrary style transfer in real-time with adaptive instance
  normalization.
\newblock In \emph{Proceedings of the IEEE international conference on computer
  vision}, pages 1501--1510, 2017.

\bibitem[Hyv{\"a}rinen and Dayan(2005)]{hyvarinen2005estimation}
Aapo Hyv{\"a}rinen and Peter Dayan.
\newblock Estimation of non-normalized statistical models by score matching.
\newblock \emph{Journal of Machine Learning Research}, 6\penalty0 (4), 2005.

\bibitem[Izmailov et~al.(2020)Izmailov, Maddox, Kirichenko, Garipov, Vetrov,
  and Wilson]{izmailov2020subspace}
Pavel Izmailov, Wesley~J Maddox, Polina Kirichenko, Timur Garipov, Dmitry
  Vetrov, and Andrew~Gordon Wilson.
\newblock Subspace inference for bayesian deep learning.
\newblock In \emph{Uncertainty in Artificial Intelligence}, pages 1169--1179.
  PMLR, 2020.

\bibitem[Karras et~al.(2017)Karras, Aila, Laine, and
  Lehtinen]{karras2017progressive}
Tero Karras, Timo Aila, Samuli Laine, and Jaakko Lehtinen.
\newblock Progressive growing of gans for improved quality, stability, and
  variation.
\newblock \emph{arXiv preprint arXiv:1710.10196}, 2017.

\bibitem[Kawar et~al.(2021)Kawar, Vaksman, and Elad]{kawar2021snips}
Bahjat Kawar, Gregory Vaksman, and Michael Elad.
\newblock Snips: Solving noisy inverse problems stochastically.
\newblock \emph{Advances in Neural Information Processing Systems},
  34:\penalty0 21757--21769, 2021.

\bibitem[Kawar et~al.(2022)Kawar, Elad, Ermon, and Song]{kawar2022denoising}
Bahjat Kawar, Michael Elad, Stefano Ermon, and Jiaming Song.
\newblock Denoising diffusion restoration models.
\newblock In \emph{Advances in Neural Information Processing Systems}, 2022.

\bibitem[Kendall and Gal(2017)]{kendall2017uncertainties}
Alex Kendall and Yarin Gal.
\newblock What uncertainties do we need in bayesian deep learning for computer
  vision?
\newblock \emph{Advances in neural information processing systems}, 30, 2017.

\bibitem[Lakshminarayanan et~al.(2017)Lakshminarayanan, Pritzel, and
  Blundell]{lakshminarayanan2017simple}
Balaji Lakshminarayanan, Alexander Pritzel, and Charles Blundell.
\newblock Simple and scalable predictive uncertainty estimation using deep
  ensembles.
\newblock \emph{Advances in neural information processing systems}, 30, 2017.

\bibitem[Li et~al.(2022)Li, Lin, Zhou, Qi, Wang, and Jia]{li2022mat}
Wenbo Li, Zhe Lin, Kun Zhou, Lu Qi, Yi Wang, and Jiaya Jia.
\newblock Mat: Mask-aware transformer for large hole image inpainting.
\newblock In \emph{Proceedings of the IEEE/CVF conference on computer vision
  and pattern recognition}, pages 10758--10768, 2022.

\bibitem[Lim et~al.(2017)Lim, Son, Kim, Nah, and Mu~Lee]{lim2017enhanced}
Bee Lim, Sanghyun Son, Heewon Kim, Seungjun Nah, and Kyoung Mu~Lee.
\newblock Enhanced deep residual networks for single image super-resolution.
\newblock In \emph{Proceedings of the IEEE conference on computer vision and
  pattern recognition workshops}, pages 136--144, 2017.

\bibitem[Liu et~al.(2015)Liu, Luo, Wang, and Tang]{liu2015faceattributes}
Ziwei Liu, Ping Luo, Xiaogang Wang, and Xiaoou Tang.
\newblock Deep learning face attributes in the wild.
\newblock In \emph{Proceedings of International Conference on Computer Vision
  (ICCV)}, 2015.

\bibitem[Louizos and Welling(2017)]{louizos2017multiplicative}
Christos Louizos and Max Welling.
\newblock Multiplicative normalizing flows for variational bayesian neural
  networks.
\newblock In \emph{International Conference on Machine Learning}, pages
  2218--2227. PMLR, 2017.

\bibitem[Lugmayr et~al.(2020)Lugmayr, Danelljan, Van~Gool, and
  Timofte]{lugmayr2020srflow}
Andreas Lugmayr, Martin Danelljan, Luc Van~Gool, and Radu Timofte.
\newblock Srflow: Learning the super-resolution space with normalizing flow.
\newblock In \emph{ECCV}, 2020.

\bibitem[Lugmayr et~al.(2022)Lugmayr, Danelljan, Romero, Yu, Timofte, and
  Van~Gool]{lugmayr2022repaint}
Andreas Lugmayr, Martin Danelljan, Andres Romero, Fisher Yu, Radu Timofte, and
  Luc Van~Gool.
\newblock Repaint: Inpainting using denoising diffusion probabilistic models.
\newblock In \emph{Proceedings of the IEEE/CVF Conference on Computer Vision
  and Pattern Recognition}, pages 11461--11471, 2022.

\bibitem[Luhman and Luhman(2021)]{luhman2021knowledge}
Eric Luhman and Troy Luhman.
\newblock Knowledge distillation in iterative generative models for improved
  sampling speed.
\newblock \emph{arXiv preprint arXiv:2101.02388}, 2021.

\bibitem[Malinin and Gales(2018)]{malinin2018predictive}
Andrey Malinin and Mark Gales.
\newblock Predictive uncertainty estimation via prior networks.
\newblock \emph{Advances in neural information processing systems}, 31, 2018.

\bibitem[Manor and Michaeli(2023)]{manor2023posterior}
Hila Manor and Tomer Michaeli.
\newblock On the posterior distribution in denoising: Application to
  uncertainty quantification.
\newblock \emph{arXiv preprint arXiv:2309.13598}, 2023.

\bibitem[Mao et~al.(2019)Mao, Lee, Tseng, Ma, and Yang]{mao2019mode}
Qi Mao, Hsin-Ying Lee, Hung-Yu Tseng, Siwei Ma, and Ming-Hsuan Yang.
\newblock Mode seeking generative adversarial networks for diverse image
  synthesis.
\newblock In \emph{Proceedings of the IEEE/CVF conference on computer vision
  and pattern recognition}, pages 1429--1437, 2019.

\bibitem[Meng et~al.(2021)Meng, Song, Li, and Ermon]{meng2021estimating}
Chenlin Meng, Yang Song, Wenzhe Li, and Stefano Ermon.
\newblock Estimating high order gradients of the data distribution by
  denoising.
\newblock \emph{Advances in Neural Information Processing Systems},
  34:\penalty0 25359--25369, 2021.

\bibitem[Meng et~al.(2022)Meng, Gao, Kingma, Ermon, Ho, and
  Salimans]{meng2022on}
Chenlin Meng, Ruiqi Gao, Diederik~P Kingma, Stefano Ermon, Jonathan Ho, and Tim
  Salimans.
\newblock On distillation of guided diffusion models.
\newblock In \emph{NeurIPS 2022 Workshop on Score-Based Methods}, 2022.

\bibitem[Metropolis et~al.(1953)Metropolis, Rosenbluth, Rosenbluth, Teller, and
  Teller]{metropolis1953equation}
Nicholas Metropolis, Arianna~W Rosenbluth, Marshall~N Rosenbluth, Augusta~H
  Teller, and Edward Teller.
\newblock Equation of state calculations by fast computing machines.
\newblock \emph{The journal of chemical physics}, 21\penalty0 (6):\penalty0
  1087--1092, 1953.

\bibitem[Monteiro et~al.(2020)Monteiro, Le~Folgoc, Coelho~de Castro, Pawlowski,
  Marques, Kamnitsas, van~der Wilk, and Glocker]{monteiro2020stochastic}
Miguel Monteiro, Lo{\"\i}c Le~Folgoc, Daniel Coelho~de Castro, Nick Pawlowski,
  Bernardo Marques, Konstantinos Kamnitsas, Mark van~der Wilk, and Ben Glocker.
\newblock Stochastic segmentation networks: Modelling spatially correlated
  aleatoric uncertainty.
\newblock \emph{Advances in neural information processing systems},
  33:\penalty0 12756--12767, 2020.

\bibitem[Neal(2012)]{neal2012bayesian}
Radford~M Neal.
\newblock \emph{Bayesian learning for neural networks}.
\newblock Springer Science \& Business Media, 2012.

\bibitem[Nehme et~al.(2023)Nehme, Yair, and Michaeli]{nehme2023uncertainty}
Elias Nehme, Omer Yair, and Tomer Michaeli.
\newblock Uncertainty quantification via neural posterior principal components.
\newblock In \emph{Thirty-seventh Conference on Neural Information Processing
  Systems}, 2023.

\bibitem[Nussbaum et~al.(2022)Nussbaum, Gawlikowski, and
  Niebling]{nussbaum2022structuring}
Frank Nussbaum, Jakob Gawlikowski, and Julia Niebling.
\newblock Structuring uncertainty for fine-grained sampling in stochastic
  segmentation networks.
\newblock \emph{Advances in Neural Information Processing Systems},
  35:\penalty0 27678--27691, 2022.

\bibitem[Ohayon et~al.(2021)Ohayon, Adrai, Vaksman, Elad, and
  Milanfar]{ohayon2021high}
Guy Ohayon, Theo Adrai, Gregory Vaksman, Michael Elad, and Peyman Milanfar.
\newblock High perceptual quality image denoising with a posterior sampling
  cgan.
\newblock In \emph{Proceedings of the IEEE/CVF International Conference on
  Computer Vision}, pages 1805--1813, 2021.

\bibitem[Ounkomol et~al.(2018)Ounkomol, Seshamani, Maleckar, Collman, and
  Johnson]{ounkomol2018label}
Chawin Ounkomol, Sharmishtaa Seshamani, Mary~M Maleckar, Forrest Collman, and
  Gregory~R Johnson.
\newblock Label-free prediction of three-dimensional fluorescence images from
  transmitted-light microscopy.
\newblock \emph{Nature methods}, 15\penalty0 (11):\penalty0 917--920, 2018.

\bibitem[Pearce et~al.(2018)Pearce, Brintrup, Zaki, and Neely]{pearce2018high}
Tim Pearce, Alexandra Brintrup, Mohamed Zaki, and Andy Neely.
\newblock High-quality prediction intervals for deep learning: A
  distribution-free, ensembled approach.
\newblock In \emph{International conference on machine learning}, pages
  4075--4084. PMLR, 2018.

\bibitem[Ritter et~al.(2018)Ritter, Botev, and Barber]{ritter2018scalable}
Hippolyt Ritter, Aleksandar Botev, and David Barber.
\newblock A scalable laplace approximation for neural networks.
\newblock In \emph{6th International Conference on Learning Representations,
  ICLR 2018-Conference Track Proceedings}. International Conference on
  Representation Learning, 2018.

\bibitem[Rivenson et~al.(2019)Rivenson, Wang, Wei, de~Haan, Zhang, Wu,
  G{\"u}nayd{\i}n, Zuckerman, Chong, Sisk, et~al.]{rivenson2019virtual}
Yair Rivenson, Hongda Wang, Zhensong Wei, Kevin de Haan, Yibo Zhang, Yichen Wu,
  Harun G{\"u}nayd{\i}n, Jonathan~E Zuckerman, Thomas Chong, Anthony~E Sisk,
  et~al.
\newblock Virtual histological staining of unlabelled tissue-autofluorescence
  images via deep learning.
\newblock \emph{Nature biomedical engineering}, 3\penalty0 (6):\penalty0
  466--477, 2019.

\bibitem[Ronneberger et~al.(2015)Ronneberger, Fischer, and
  Brox]{ronneberger2015u}
Olaf Ronneberger, Philipp Fischer, and Thomas Brox.
\newblock U-net: Convolutional networks for biomedical image segmentation.
\newblock In \emph{Medical Image Computing and Computer-Assisted
  Intervention--MICCAI 2015: 18th International Conference, Munich, Germany,
  October 5-9, 2015, Proceedings, Part III 18}, pages 234--241. Springer, 2015.

\bibitem[Russakovsky et~al.(2015)Russakovsky, Deng, Su, Krause, Satheesh, Ma,
  Huang, Karpathy, Khosla, Bernstein, and Li]{russakovsky2015imagenet}
Olga Russakovsky, Jia Deng, Hao Su, Jonathan Krause, Sanjeev Satheesh, Sean Ma,
  Zhiheng Huang, Andrej Karpathy, Aditya Khosla, Michael Bernstein, and Fei-Fei
  Li.
\newblock Imagenet large scale visual recognition challenge.
\newblock \emph{International journal of computer vision}, 115:\penalty0
  211--252, 2015.

\bibitem[Saharia et~al.(2022{\natexlab{a}})Saharia, Chan, Chang, Lee, Ho,
  Salimans, Fleet, and Norouzi]{saharia2022palette}
Chitwan Saharia, William Chan, Huiwen Chang, Chris Lee, Jonathan Ho, Tim
  Salimans, David Fleet, and Mohammad Norouzi.
\newblock Palette: Image-to-image diffusion models.
\newblock In \emph{ACM SIGGRAPH 2022 Conference Proceedings}, pages 1--10,
  2022{\natexlab{a}}.

\bibitem[Saharia et~al.(2022{\natexlab{b}})Saharia, Ho, Chan, Salimans, Fleet,
  and Norouzi]{saharia2022image}
Chitwan Saharia, Jonathan Ho, William Chan, Tim Salimans, David~J Fleet, and
  Mohammad Norouzi.
\newblock Image super-resolution via iterative refinement.
\newblock \emph{IEEE Transactions on Pattern Analysis and Machine
  Intelligence}, 2022{\natexlab{b}}.

\bibitem[Salimans and Ho(2022)]{salimans2022progressive}
Tim Salimans and Jonathan Ho.
\newblock Progressive distillation for fast sampling of diffusion models.
\newblock \emph{arXiv preprint arXiv:2202.00512}, 2022.

\bibitem[Salimans et~al.(2015)Salimans, Kingma, and
  Welling]{salimans2015markov}
Tim Salimans, Diederik Kingma, and Max Welling.
\newblock Markov chain monte carlo and variational inference: Bridging the gap.
\newblock In \emph{International conference on machine learning}, pages
  1218--1226. PMLR, 2015.

\bibitem[Sankaranarayanan et~al.(2022)Sankaranarayanan, Angelopoulos, Bates,
  Romano, and Isola]{sankaranarayanansemantic}
Swami Sankaranarayanan, Anastasios~Nikolas Angelopoulos, Stephen Bates, Yaniv
  Romano, and Phillip Isola.
\newblock Semantic uncertainty intervals for disentangled latent spaces.
\newblock In \emph{Advances in Neural Information Processing Systems}, 2022.

\bibitem[Sehwag et~al.(2022)Sehwag, Hazirbas, Gordo, Ozgenel, and
  Canton]{sehwag2022generating}
Vikash Sehwag, Caner Hazirbas, Albert Gordo, Firat Ozgenel, and Cristian
  Canton.
\newblock Generating high fidelity data from low-density regions using
  diffusion models.
\newblock In \emph{Proceedings of the IEEE/CVF Conference on Computer Vision
  and Pattern Recognition}, pages 11492--11501, 2022.

\bibitem[Sohl-Dickstein et~al.(2015)Sohl-Dickstein, Weiss, Maheswaranathan, and
  Ganguli]{sohl2015deep}
Jascha Sohl-Dickstein, Eric Weiss, Niru Maheswaranathan, and Surya Ganguli.
\newblock Deep unsupervised learning using nonequilibrium thermodynamics.
\newblock In \emph{International conference on machine learning}, pages
  2256--2265. PMLR, 2015.

\bibitem[Song and Ermon(2019)]{song2019generative}
Yang Song and Stefano Ermon.
\newblock Generative modeling by estimating gradients of the data distribution.
\newblock \emph{Advances in neural information processing systems}, 32, 2019.

\bibitem[Song et~al.(2020)Song, Sohl-Dickstein, Kingma, Kumar, Ermon, and
  Poole]{song2020score}
Yang Song, Jascha Sohl-Dickstein, Diederik~P Kingma, Abhishek Kumar, Stefano
  Ermon, and Ben Poole.
\newblock Score-based generative modeling through stochastic differential
  equations.
\newblock \emph{arXiv preprint arXiv:2011.13456}, 2020.

\bibitem[Song et~al.(2021{\natexlab{a}})Song, Durkan, Murray, and
  Ermon]{song2021maximum}
Yang Song, Conor Durkan, Iain Murray, and Stefano Ermon.
\newblock Maximum likelihood training of score-based diffusion models.
\newblock \emph{Advances in Neural Information Processing Systems},
  34:\penalty0 1415--1428, 2021{\natexlab{a}}.

\bibitem[Song et~al.(2021{\natexlab{b}})Song, Shen, Xing, and
  Ermon]{song2021solving}
Yang Song, Liyue Shen, Lei Xing, and Stefano Ermon.
\newblock Solving inverse problems in medical imaging with score-based
  generative models.
\newblock In \emph{International Conference on Learning Representations},
  2021{\natexlab{b}}.

\bibitem[Song et~al.(2023)Song, Dhariwal, Chen, and
  Sutskever]{song2023consistency}
Yang Song, Prafulla Dhariwal, Mark Chen, and Ilya Sutskever.
\newblock Consistency models.
\newblock \emph{arXiv preprint arXiv:2303.01469}, 2023.

\bibitem[Sutskever and Tieleman(2010)]{sutskever2010convergence}
Ilya Sutskever and Tijmen Tieleman.
\newblock On the convergence properties of contrastive divergence.
\newblock In \emph{Proceedings of the thirteenth international conference on
  artificial intelligence and statistics}, pages 789--795. JMLR Workshop and
  Conference Proceedings, 2010.

\bibitem[von Chamier et~al.(2021)von Chamier, Laine, Jukkala, Spahn, Krentzel,
  Nehme, Lerche, Hern{\'a}ndez-P{\'e}rez, Mattila, Karinou,
  et~al.]{von2021democratising}
Lucas von Chamier, Romain~F Laine, Johanna Jukkala, Christoph Spahn, Daniel
  Krentzel, Elias Nehme, Martina Lerche, Sara Hern{\'a}ndez-P{\'e}rez, Pieta~K
  Mattila, Eleni Karinou, et~al.
\newblock Democratising deep learning for microscopy with zerocostdl4mic.
\newblock \emph{Nature communications}, 12\penalty0 (1):\penalty0 2276, 2021.

\bibitem[Wang et~al.(2019)Wang, Li, Aertsen, Deprest, Ourselin, and
  Vercauteren]{wang2019aleatoric}
Guotai Wang, Wenqi Li, Michael Aertsen, Jan Deprest, S{\'e}bastien Ourselin,
  and Tom Vercauteren.
\newblock Aleatoric uncertainty estimation with test-time augmentation for
  medical image segmentation with convolutional neural networks.
\newblock \emph{Neurocomputing}, 338:\penalty0 34--45, 2019.

\bibitem[Wang et~al.(2023)Wang, Yu, and Zhang]{wang2023ddnm}
Yinhuai Wang, Jiwen Yu, and Jian Zhang.
\newblock Zero-shot image restoration using denoising diffusion null-space
  model.
\newblock \emph{The Eleventh International Conference on Learning
  Representations}, 2023.

\bibitem[Welling and Teh(2011)]{welling2011bayesian}
Max Welling and Yee~W Teh.
\newblock Bayesian learning via stochastic gradient langevin dynamics.
\newblock In \emph{Proceedings of the 28th international conference on machine
  learning (ICML-11)}, pages 681--688, 2011.

\bibitem[Yair and Michaeli(2022)]{yair2022thinking}
Omer Yair and Tomer Michaeli.
\newblock Thinking fourth dimensionally: Treating time as a random variable in
  ebms.
\newblock \emph{https://openreview.net/forum?id=m0fEJ2bvwpw}, 2022.

\bibitem[Yu et~al.(2020)Yu, Li, Zhou, Malik, Davis, and Fritz]{yu2020inclusive}
Ning Yu, Ke Li, Peng Zhou, Jitendra Malik, Larry Davis, and Mario Fritz.
\newblock Inclusive gan: Improving data and minority coverage in generative
  models.
\newblock In \emph{Computer Vision--ECCV 2020: 16th European Conference,
  Glasgow, UK, August 23--28, 2020, Proceedings, Part XXII 16}, pages 377--393.
  Springer, 2020.

\bibitem[Zhao et~al.(2020)Zhao, Xie, and Li]{zhao2020learning}
Yang Zhao, Jianwen Xie, and Ping Li.
\newblock Learning energy-based generative models via coarse-to-fine expanding
  and sampling.
\newblock In \emph{International Conference on Learning Representations}, 2020.

\end{thebibliography}
}


\end{document}